\documentclass{article}

\PassOptionsToPackage{numbers, compress}{natbib}

\usepackage[preprint]{neurips_2025}

\usepackage[utf8]{inputenc} % allow utf-8 input
\usepackage[T1]{fontenc}    % use 8-bit T1 fonts
\usepackage{hyperref}       % hyperlinks
\usepackage{url}            % simple URL typesetting
\usepackage{booktabs}       % professional-quality tables
\usepackage{amsfonts}       % blackboard math symbols
\usepackage{nicefrac}       % compact symbols for 1/2, etc.
\usepackage{microtype}      % microtypography
\usepackage{xcolor}         % colors

\usepackage{multirow}
\usepackage{float}
\usepackage{graphicx}
\usepackage{array}
\usepackage{amsmath}
\usepackage{amssymb}
\usepackage{amsthm}
\usepackage{cleveref}
\usepackage{wrapfig}
\usepackage{caption}
\usepackage{longtable}
\usepackage{makecell}
\usepackage{enumitem}

\newtheorem{theorem}{Theorem} 

\title{Adaptive Task Vectors for Large Language Models}

\author{
    Joonseong Kang$^*$ \quad Soojeong Lee \quad Subeen Park \quad Sumin Park \\
    \textbf{Taero Kim} \quad \textbf{Jihee Kim} \quad \textbf{Ryunyi Lee} \quad \textbf{Kyungwoo Song}$^\dagger$ \\\\
    Yonsei University
}

\begin{document}

\maketitle

\def\thefootnote{*}\footnotetext{doongsae@yonsei.ac.kr, $^\dagger$Corresponding author: kyungwoo.song@yonsei.ac.kr}\def\thefootnote{\arabic{footnote}}

\begin{abstract}

In-Context Learning (ICL) enables Large Language Models (LLMs) to perform tasks without parameter updates by conditioning on a few demonstrations provided in the prompt. Despite its success, ICL suffers from several limitations, including sensitivity to demonstration order, context length constraints, and computational inefficiency. 
To address these challenges, task vector-based approaches compress task information into a single vector. However, these methods typically construct task vectors from fixed sets of demonstrations and reuse them across input queries, without conditioning on the specific input. This limitation can lead models to struggle with effective adaptation when the input query is not well aligned with the underlying demonstrations, consequently degrading their generalization performance on unseen tasks.
To overcome this limitation, we propose \textbf{Adaptive Task Vectors (ATV)}, a simple and effective framework that dynamically generates task vectors conditioned on each input query. ATV employs a small language model to generate task vectors, which are then transformed to match the target LLM’s architecture and applied to guide its output generation. 
In contrast to ICL and previous vector-based approaches, which rely on fixed demonstration sets and their corresponding vectors, ATV dynamically generates task vectors tailored to each specific input query and task. Consequently, ATV demonstrates strong performance and generalization capabilities, even for unseen tasks.
Furthermore, we provide a theoretical analysis indicating that ATV is expressively equivalent to LoRA under equal rank budgets and more expressive than Prefix-Tuning, thereby offering formal support for its representational advantage.
\end{abstract}

\section{Introduction}
\label{introduction}

Large Language Models (LLMs) have made remarkable progress in natural language processing, demonstrating impressive performance across various tasks. In-Context Learning (ICL)~\cite{brown2020language} has become a pivotal method for enhancing LLM performance, enabling models to effectively perform specific tasks by including demonstration samples in prompts without requiring additional training~\cite{dong2022survey}.
However, ICL faces several limitations: performance varies considerably depending on the order and selection of demonstration samples~\cite{liu2021makes, peng2024revisiting, wang2023large}, the maximum context length constraint of LLMs makes it challenging to handle tasks involving long-context reasoning or diverse demonstration sets, and processing numerous demonstration significantly reduces computational efficiency~\cite{li2024long, kuratov2024babilong}.

To mitigate these issues, task vector-based approaches~\cite{hendel-etal-2023-context, ilharco2023editing, liu2024context, li2025implicit, wang2025elicit} have attracted growing interest for improving the efficiency and robustness of ICL.
Task vectors~\cite{hendel-etal-2023-context} are vector representations that compress task-specific information, typically obtained from the hidden state of the last token in the prompt, or its processed variant. These vectors are integrated with the input query to modulate the model’s output in a task-specific manner.
Recent studies have utilized task vectors to effectively mitigate the limitations of conventional ICL~\cite{yang2025task, huang2024multimodal}. By compressing information from multiple demonstration samples into a single vector, these methods overcome context window constraints and reduce performance variability due to demonstration ordering~\cite{dong2022survey, zhao2021calibrate, lu2021fantastically, zhang2024batch}. As a result, they preserve the effectiveness of ICL while improving computational efficiency and consistency~\cite{li2025task}.

However, existing task vector-based approaches exhibit a significant limitation.
Most prior methods construct task vectors from fixed sets of demonstration samples and reuse the same vector across all input queries, regardless of their individual characteristics ~\cite{hendel-etal-2023-context, yang2025task}.
While some recent approaches retrieve task vectors based on query similarity, the retrieved vectors are precomputed and remain fixed during inference. As a result, these methods are not conditioned on the current input and may fail to adapt effectively when the input is not well aligned with the underlying demonstrations. Indeed, ICL and previous task vector-based methods, which select demonstration sets from such fixed pools, consequently tend to exhibit limited performance on unseen tasks.

\begin{figure}[t]
    \centering
    \includegraphics[width=\textwidth]{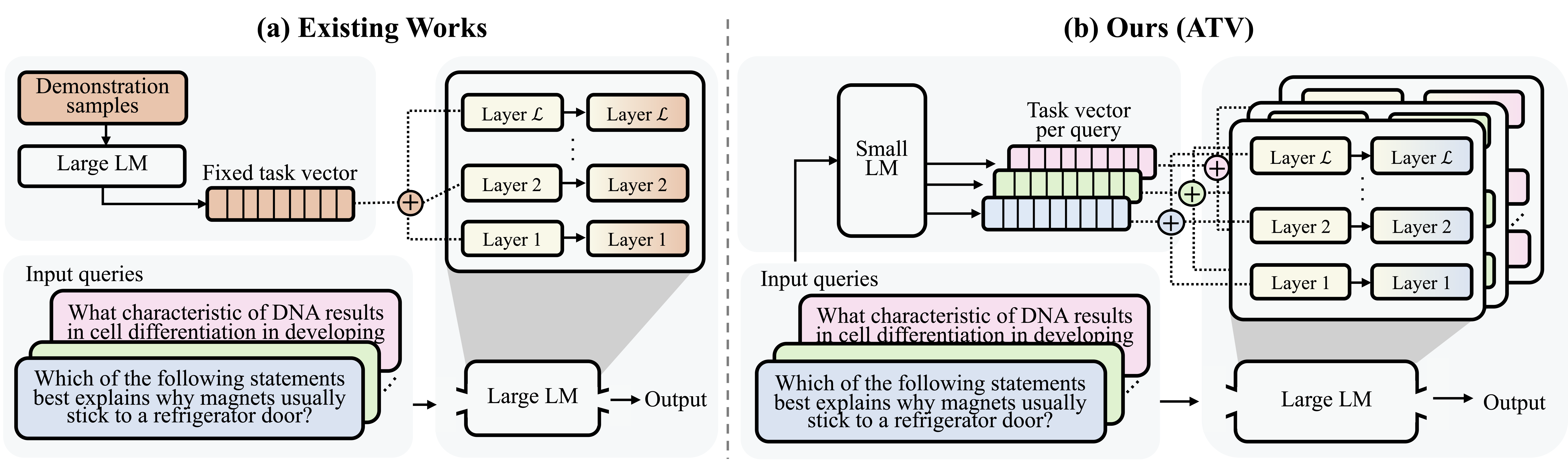}
    \caption{Comparison between task vector methods: a) Prior work uses a fixed task vector for all inputs, whereas b) our method generates a query-specific task vector, enabling adaptive behavior for each input. This enables the LLM to adapt its behavior to individual inputs and overcome the limitations of fixed-vector approaches.}
    \vspace{-0.5em}
    \label{fig:fig1_intro}
\end{figure}

Motivated by these limitations, we present \textbf{Adaptive Task Vectors (ATV)}, a new framework for dynamically generating task vectors conditioned on each input query. ATV enables more accurate and input-sensitive model guidance by producing an optimal task vector for each input query. Our framework employs a small language model to generate intermediate task representations, which are then transformed to match the architecture of the target LLM and used to modulate its output. Figure~\ref{fig:fig1_intro} illustrates the key difference between (a) conventional fixed-vector methods~\cite{liu2024context,li2025implicit,wang2025elicit}, and (b) our adaptive approach. While (a) applies the same vector to all input queries, (b) generates a query-specific vector for each query, allowing the model to produce more appropriate responses tailored to the input.

In this paper, we establish the effectiveness of the proposed framework through both theoretical and empirical evaluation. Theoretically, we prove that ATV is expressively equivalent to LoRA under matched rank budgets and strictly more expressive than Prefix-Tuning, offering a formal basis for its enhanced representational capacity. Empirically, we evaluate ATV on in-domain performance, generalization to unseen tasks, and ablations on model capacity and injection configuration. Across these settings, ATV demonstrates superior in-domain accuracy, strong generalization ability, and interpretable insights into model capacity and injection behavior.

\paragraph{Our main contributions are as follows:}
\begin{itemize}
    \item We propose \textbf{Adaptive Task Vectors (ATV)}, a simple and effective framework that generates task vectors conditioned on each input query, enabling LLMs to adapt their behavior in a task-aware manner based on the input.
    \item We provide a theoretical analysis showing that ATV is expressively equivalent to LoRA under equal rank budgets and strictly more expressive than Prefix-Tuning, providing a formal justification for its enhanced representational capacity.
    \item We empirically evaluate ATV on both in-domain tasks and generalization to unseen tasks, demonstrating strong performance across diverse datasets and model families. We further analyze how ATV’s performance and behavior are influenced by key design factors through ablation studies on model capacity and injection configuration.
\end{itemize}

\section{Related Work}
\label{related_work}

\paragraph{In-Context Learning.}

ICL enables LLMs to perform new tasks without parameter updates by conditioning on a few input-output examples presented in the prompt~\cite{brown2020language}. Since the introduction of GPT-3, ICL has demonstrated strong performance across diverse tasks, particularly when combined with prompt engineering and model scaling~\cite{wei2022chain, kojima2022large, zhou2022least}. Despite its effectiveness, ICL exhibits notable limitations. First, its performance is highly sensitive to the order and selection of demonstration samples~\cite{zhao2021calibrate, min2022rethinking}. Second, it is constrained by the maximum context length of LLMs, limiting the complexity and coverage of tasks~\cite{li2024long}. Third, processing multiple demonstrations during inference incurs significant computational costs~\cite{kuratov2024babilong}.  

To mitigate these issues, recent works have explored adaptive or retrieval-based ICL, which dynamically selects demonstration examples based on the input query~\cite{rubin2022learning}. While such methods mitigate order sensitivity, they still rely on in-context tokens and remain constrained by context length. More fundamentally, ICL depends on explicit prompt tokens to convey task information. 
In contrast, our method departs from token-based prompting by introducing a compact, learned vector that conveys task information without explicit demonstrations. This approach retains ICL’s key strengths, such as query-specific adaptation and task generalization, while eliminating challenges related to prompt design and input length.

\paragraph{Vector-Based Approaches for Model Steering.}
Recent work has explored replacing in-context demonstrations with task vectors, which are dense representations that encode task information derived from a few-shot prompt. These vectors are typically extracted from the hidden state of the last token of the demonstration prompt within a transformer model~\cite{hendel-etal-2023-context}, effectively summarizing the task’s semantics in a single activation. For instance, Implicit In-Context Learning (I2CL)~\cite{li2025implicit} compresses demonstration examples into a single context vector and injects it into the model’s residual stream via linear operations. Similarly, ELICIT~\cite{wang2025elicit} maintains a library of task vectors associated with specific capabilities and retrieves the appropriate one based on the input.

These methods improve efficiency by removing token-level demonstrations. However, the task vectors they rely on are either derived from a fixed set of demonstrations or retrieved from a static library, and are reused across all input queries. As a result, they are not conditioned on the input and offer limited adaptability at inference time.
To our knowledge, existing methods do not support the generation of task vectors conditioned on each input. Our proposed framework, ATV, addresses this limitation by dynamically generating task vectors per query, enabling more fine-grained input-aware task representation. This combines the efficiency of vector-based approaches with the adaptability required for input-level variation.

\section{Methodology}
\label{methodology}

\subsection{Background and Preliminaries}
\label{sec:background}

\paragraph{Transformer Architecture.}
The Transformer is a neural architecture based on self-attention and forms the foundation of modern large-scale language models~\cite{vaswani2017attention}. Each layer consists of a feed-forward network and a self-attention mechanism, which allows tokens to attend to others in the sequence. The self-attention operation is defined as:
\begin{equation}
    \text{Attn}(Q, K, V) = \text{softmax}\left(\frac{QK^T}{\sqrt{d_k}}\right)V
    \label{eq:attention}
\end{equation}
where $Q$, $K$, and $V$ indicate the query, key, and value matrices, respectively, and $d_k$ is the dimensionality of the key vectors.

Let $h_t^{l-1}$ be the hidden state of token $t$ at layer $l{-}1$. A standard Transformer layer updates it as:
\begin{align}
    \tilde{h}_t^l &= \text{LayerNorm}\left(h_t^{l-1} + \text{Attn}(Q_t, K, V)\right) \\
    h_t^l &= \text{LayerNorm}\left(\tilde{h}_t^l + \text{MLP}(\tilde{h}_t^l)\right)
    \label{eq:attn_update}
\end{align}
where $Q_t = W_Q h_t^{l-1}$ and $K, V$ are projections of the previous layer’s hidden states.
In auto-regressive models, the hidden state of the last token $h_T^l$ summarizes the input context and is used for next-token prediction.

\paragraph{Task Vectors.}
Following prior work~\cite{wang2025elicit}, we define a \textit{task vector} as the hidden state of the last token at each transformer layer, capturing task-relevant information in a compressed form. Given an input $x = [x_1, x_2, \dots, x_T]$, the task vector at layer $l$ is:
\begin{equation}
v_{\text{task}}^l = h_T^l
\quad \text{(task vector extracted from the last token at layer $l$)}
\label{eq:task-vector}
\end{equation}

To steer the model output in a task-specific direction, we inject the task vector into the hidden state of the last token at each transformer layer. Specifically, for each layer \( l \), the modified hidden state is computed as:
\begin{equation}
    \tilde{h}^l = h^l + \lambda v_{\text{task}}^l
\end{equation}
where \( h^l \) denotes the hidden state of the last token at layer \( l \), \( v_{\text{task}}^l \in \mathbb{R}^{d_l} \) is the corresponding task vector slice, \( \tilde{h}^l \) is the injected version, \( d_l \) is the hidden dimensionality at layer \( l \), and \( \lambda \) is a scaling factor controlling the strength of the intervention.
For simplicity, we omit the token index and refer to the last token's hidden state simply as \( h^l \). This formulation allows the task vector to modulate the model’s behavior in a lightweight and interpretable manner.

Previous methods rely on task vectors extracted from fixed demonstrations, resulting in a static representation shared across inputs. We introduce the \textbf{Adaptive Task Vector (ATV)}, which is dynamically generated per input to modulate the model’s behavior.

\subsection{ATV: Adaptive Task Vectors for Large Language Models}
\label{sec:method}

\paragraph{Notation and Setup.}
Let \( x = [x_1, x_2, \dots, x_T] \) be a tokenized input query sequence of length \( T \), and let \( y \) denote the corresponding target output. We define two models:
a small model \( \mathcal{M}_{\text{small}} \) with hidden size \( d_s \), and a large language model \( \mathcal{M}_{\text{large}} \) with \( L \) layers and hidden size \( d_l \) per layer.

From the input \( x \), \( \mathcal{M}_{\text{small}} \) produces a hidden representation \( v_{\text{small}} \in \mathbb{R}^{d_s} \), extracted from the last token of the last layer. This vector is then expanded via a parameterized function \( f_\theta: \mathbb{R}^{d_s} \rightarrow \mathbb{R}^{L \times d_l} \) to obtain our \textbf{ATV} \( v_{\text{ATV}} = f_\theta(v_{\text{small}}) \), suitable for injection into the large model. Let \( v_{\text{ATV}}^l \in \mathbb{R}^{d_l} \) denote the portion corresponding to the \( l \)-th layer. The final output generated by \( \mathcal{M}_{\text{large}} \) after ATV injection is denoted \( \tilde{y} \). The ATV is scaled by a hyperparameter \( \lambda \) before being added to the hidden state.

\paragraph{Overview of the ATV Framework.}
Our goal is to steer an LLM without modifying its weights by injecting input-conditioned signals directly into its hidden states. We introduce \textbf{ATV}, a lightweight control framework that operates externally to a frozen large model.

ATV consists of two lightweight modules:

(1) \textbf{ATV generation.} A small language model produces a compact vector representation from the input query.

(2) \textbf{ATV expansion.} An expansion module transforms this vector into a set of layer-wise steering signals injected into the large model.

We implement the expansion module as a single linear projection from \( \mathbb{R}^{d_s} \) to \( \mathbb{R}^{L \cdot d_l} \), followed by reshaping into \( \mathbb{R}^{L \times d_l} \) for compatibility with the target model.  
The generator and expansion modules are trained jointly, while the LLM remains frozen.

By injecting the ATV into the internal layers of the large model, the ATV enables flexible and targeted control over the model's behavior. This allows the large model to better align with desired task objectives, such as answering questions accurately or performing structured reasoning, without modifying its parameters or relying on prompt engineering.

\begin{figure}[t]
  \centering
  \includegraphics[width=\linewidth]{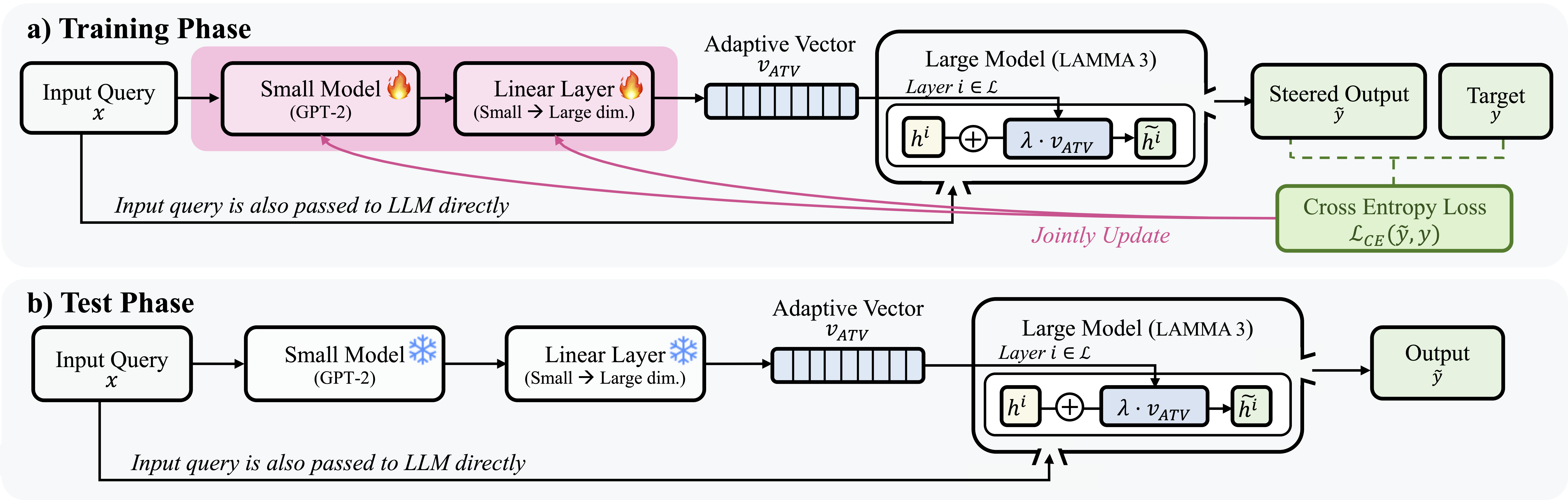}
  \caption{Overview of the Adaptive Task Vector (ATV) framework. Top: During training, the small model and expansion module are updated to minimize the loss from the steered output. Bottom: During inference, both modules are frozen and used to generate query-specific ATV vectors for steering the frozen large model.}
  \label{fig:framework}
  \vspace{-10pt}
\end{figure}

We summarize this process in Figure~\ref{fig:framework}. During training (top), the small model and the expansion module are optimized to produce effective ATVs that steer the large model toward the desired output. During inference (bottom), both modules are frozen and used to dynamically generate ATVs for new input queries.

The core idea behind ATV is to adapt the behavior of an LLM to a specific task without modifying its parameters. Instead of prompt-based conditioning, we steer the model by injecting query-specific information directly into its internal hidden states in the form of an ATV.

To generate the ATV, we first encode the input query using a small language model \( \mathcal{M}_{\text{small}} \), such as a GPT-2 variant~\cite{radford2019language}. The model processes the full tokenized sequence and produces hidden representations at each layer. From the last token of the last layer, we extract a compact vector \( v_{\text{small}} \) that summarizes the semantics of the input query.

This vector is then expanded by \( f_\theta \) into the \textbf{ATV} \( v_{\text{ATV}} \), where each slice \( v_{\text{ATV}}^l \) is designed to modulate computations at the corresponding layer of the large model.

The ATV is injected into the frozen large model by modifying the hidden states of the \textit{last token} during the forward pass. Specifically, for each layer \( l \), the original hidden state \( h^l \) of the last token is modified as:
\begin{equation}
    \tilde{h}^l = h^l + \lambda v_{\text{ATV}}^l
    \label{eq:injection}
\end{equation}
where \( \lambda \) is a scalar hyperparameter that scales the ATV's influence. This additive injection provides lightweight yet effective steering of the model's output behavior.

To enable effective learning, we jointly train \( \mathcal{M}_{\text{small}} \) and \( f_\theta \) using a supervised loss. Let \( y \) denote the target output for input query \( x \), and let \( \tilde{y} \) be the output of the large model after ATV injection. The objective is to minimize the cross-entropy loss:
\begin{equation}
    \min_{\phi, \theta} \; \mathbb{E}_{(x, y) \sim \mathcal{D}} \left[ \mathcal{L}_{\text{CE}}\left(\tilde{y}, \; y \right) \right], \quad \text{where } \tilde{y} = \mathcal{M}_{\text{large}}(x; v_{\text{ATV}})
    \label{eq:loss}
\end{equation}
where \( \mathcal{D} \) denotes the supervised training dataset, and \( \phi \), \( \theta \) are the parameters of the small model and the expansion module, respectively. Notably, the large model \( \mathcal{M}_{\text{large}} \) remains frozen throughout training.

After training, both \( \mathcal{M}_{\text{small}} \) and \( f_\theta \) are frozen. During inference, given a new input query, the system generates a query-specific ATV and injects it into the large model to guide its behavior. This enables ATV to adapt frozen LLMs to diverse downstream tasks in a modular and parameter-efficient manner.
\subsection{Theoretical Analysis}
\label{sec:theoretical_analysis}
We now theoretically analyze the proposed ATV framework to better understand its effect on the behavior of the LLM by comparing it to two prominent parameter-efficient tuning methods: LoRA and Prefix-Tuning. LoRA (Low-Rank Adaptation) injects trainable low-rank matrices into pretrained weights and has demonstrated strong performance in language model fine-tuning~\cite{hu2022lora}.
Prefix-Tuning prepends trainable continuous vectors to the input sequence, conditioning the model through modified attention mechanisms without altering the pretrained weights~\cite{li2021prefix}.

Specifically, we focus on addressing the following two questions: (1) How does ATV compare to LoRA in expressivity under the same rank budget, and when might ATV offer additional advantages? (2) Does ATV offer a more expressive attention mechanism compared to Prefix-Tuning?

We first show that ATV is never weaker than LoRA when both methods operate under the same rank budget, and then clarify in which respects ATV can be stronger. 
\begin{theorem}(ATV-LoRA equivalence under equal rank)\label{thm:atv_lora}
% \paragraph{Theorem 1 (ATV--LoRA equivalence under equal rank)} 
Let $h^\ell \in \mathbb{R}^{d_\ell}$ be the last-token hidden state at layer $\ell$, and let $\tilde{h}^\ell$, $\hat{h}^\ell$ be the outputs of ATV and LoRA. For $d_s \ll d_\ell$ and a LoRA rank $r = d_s$, ATV and LoRA are expressively equivalent: for any ATV update $\tilde{h}^\ell$, there exists a LoRA configuration yielding $\hat{h}^\ell = \tilde{h}^\ell$, and vice versa (see Appendix~\ref{app:proof_thrm1} for the full proof).
\end{theorem}

\paragraph{Situations in which ATV exceeds LoRA}
The equivalence theorem~\ref{thm:atv_lora} ensures that ATV inherits LoRA's expressiveness. Beyond this, there are several aspects in which ATV can exceed LoRA. First, if the auxiliary bottleneck is widened ($d_s > r$), ATV can represent update directions that LoRA cannot capture within its fixed rank-$r$ subspace, thereby increasing its expressive power and enhancing its flexibility to adapt across diverse tasks and input distributions.
Second, even with $d_s=r$, ATV enables effective direction injection as it induces input-dependent perturbations. In the ATV-to-LoRA construction, the matrix that plays the role of LoRA's down-projection matrix $W_{down}$, a factor that remains static during inference, is given by $M := x^{+\top}(\lambda v_{small})$, which, unlike LoRA, varies with the current activation $x$. Here, $\lambda$ is a scalar hyperparmeter of ATV and $x^{+\top}$ denotes the Moore-Penrose pseudoinverse of $x^{\top}$ (see Step 3 of the proof in Appendix ~\ref{app:proof_thrm1_proof}).
This query dependence allows ATV to adapt its update on a per-instance basis, an ability LoRA lacks, which may improve adaptability in dynamic or multi-task environments. 

Secondly, under the relaxed linear attention approximation, we argue in Theorem ~\ref{thm:atv_vs_prefix} that any representation obtainable by prefix-tuning is also realizable by ATV, while the converse does not hold. 
To examine the source of expressivity differences under the approximation proposed by prior work~\cite{dai2022can}, we begin by formulating the standard attention as $\text{Attn}(xW_q, CW_k, CW_v)$, where $x \in \mathbb{R}^{T \times d_l}$ is the query, $C \in \mathbb{R}^{m \times d_l}$ is the length-$m$ context, and $W_q, W_k, W_v$ are projection matrices.
Prefix-tuning modifies the key and value by concatenating $p$ trainable prefix vectors $P_k, P_v \in \mathbb{R}^{p \times d_l}$, yielding an augmented attention~\cite{he2022towards}:$\text{Attn}_{\text{prefix}} = \text{Attn}(x W_q,[P_k; CW_k], [P_v; CW_v])$, ATV, in contrast, injects a trained vector $v_{ATV}^l$ additively to both the query and context: $\text{Attn}_{\text{ATV}}= \text{Attn}\left((x+e_{T} \cdot ({v_{ATV}^l})^\top )W_q,(C+e_{m} \cdot ({v_{ATV}^l})^\top)W_k,(C+e_{m} \cdot ({v_{ATV}^l})^\top)W_v\right)$, where $e_{m}$is a vector $ [0, ... 0, 1] \in \mathbb{R}^{m \times 1}$. 
\begin{theorem} (ATV is more expressive than Prefix-Tuning) \label{thm:atv_vs_prefix}
Let $\text{Attn}_{\text{ATV}}$ and $\text{Attn}_{\text{prefix}}$ denote the attention outputs from ATV and prefix-tuning, respectively. Then, the representational space $\mathcal{F}$ of $\text{Attn}_{\text{ATV}}$ includes that of $\text{Attn}_{\text{prefix}}$:
\begin{equation}
\mathcal{F}(\text{Attn}_{\text{prefix}}) \subseteq \mathcal{F}(\text{Attn}_{\text{ATV}})
\end{equation}
\end{theorem}
\paragraph{Comparison of $\text{Attn}_{\text{ATV}}$ and $\text{Attn}_{\text{prefix}}$}%Attention Output: Prefix-Tuning vs. ATV
Under the approximation, $\text{Attn}_{\text{ATV}} \approx \text{Attn}_{\text{prefix}} + \Delta_\text{cross}$, where $\Delta_{\text{cross}}$ encapsulates the six cross-terms that capture the additional interactions between the query and context, modulated by the vector $v^l_{\text{ATV}}$ (see Appendix~\ref{app:proof_thrm2} for the full proof).

\section{Experiments}
\label{sec:experiments}
To evaluate ATV, we design experiments examining both in-domain and generalization performance to unseen tasks, followed by ablation studies on the generator model capacity and injection strategies. We also compare ATV with parameter-efficient tuning methods to validate our theoretical analysis and visualize task vector distributions to understand their representational properties.
We closely follow ELICIT's~\cite{wang2025elicit} experimental design, using identical datasets and evaluation protocols. 
We evaluate on LLaMA3-8B~\cite{grattafiori2024llama} and Mistral-7B~\cite{jiang2023mistral7b}, with I2CL~\cite{li2025implicit} as an additional baseline and a separate comparison to LoRA~\cite{hu2022lora} for theoretical validation.
Our code is available at \url{https://github.com/MLAI-Yonsei/ATV}.

\subsection{Experiment Setup}
\paragraph{\textbf{Models and Baselines.}}
Our primary models are LLaMA3-8B and Mistral-7B.
We compare ATV against (i) zero-shot, (ii) 16-shot in-context learning (ICL), (iii) 16-shot BM25 retrieval-based ICL~\cite{robertson2009probabilistic}, (iv) ELICIT~\cite{wang2025elicit}, and (v) I2CL~\cite{li2025implicit}.
ICL and BM25 baselines use 16 demonstrations, either randomly sampled or retrieved from the task’s training set.

\paragraph{\textbf{Datasets.}}
We evaluate ATV on a diverse collection of 20 tasks spanning five categories: Knowledge, Reasoning, Mathematics, Safety, and Natural Language Understanding.
These tasks assess various NLP capabilities from reasoning to numerical problem solving.
In addition to in-domain tasks, we evaluate ATV on a separate set of unseen tasks to assess its generalization ability.

\paragraph{\textbf{Evaluation.}}
We adopt the same evaluation strategy as ELICIT to reflect realistic inference scenarios, where test-time query formats differ from those seen during training. 
Each test query is presented in two additional template variations not used in task vector generation. 
Task-specific instructions are prepended for all methods to ensure fair comparison.
Full implementation and experimental details, including dataset names, template formats, and hyperparameters, are available in Appendix~\ref{app:exp_setting}.

\subsection{In-Domain Performance Evaluation}
\label{sec:in-domain}
\vspace{-10pt}
\begin{table}[th]
    \caption{
    \textbf{In-domain performance comparison across five categories under LLaMA3 and Mistral.}
    ATV achieves the highest average accuracy on both models using the same number of tokens as ELICIT and I2CL, while outperforming all baselines across most domains and maintaining superior token efficiency over prompt-based methods.
    All results except I2CL and ATV are from ELICIT~\cite{wang2025elicit}.
    }
    \vspace{0.5em}
    \label{tab:tab_1_main_exp}
    \centering
    \small
    \renewcommand{\arraystretch}{1.05}
    \resizebox{0.95\textwidth}{!}{%
    \begin{tabular}{cc|c|ccccc|c}
        \toprule
        \multicolumn{2}{c|}{\textbf{Model}} & \textbf{\# Tokens} & \textbf{NLU} & \textbf{Reasoning} & \textbf{Knowledge} & \textbf{Math} & \textbf{Safety} & \textbf{Avg.} \\ 
        \midrule
        \multirow{6}{*}{\textbf{Llama3}}  
        & Zero-shot       & 108.3 $\pm$ 1.4 & 32.2 $\pm$ 1.2 & 31.6 $\pm$ 0.2 & 42.5 $\pm$ 1.2 & 14.0 $\pm$ 1.1 & 35.5 $\pm$ 1.2 & 31.1 $\pm$ 1.0 \\
        & 16-shot         & 1883.8 $\pm$ 0.9 & \underline{60.6 $\pm$ 0.9} & 56.0 $\pm$ 0.4 & \underline{70.6 $\pm$ 1.0} & \underline{26.7 $\pm$ 2.0} & \underline{62.1 $\pm$ 0.3} & \underline{55.2 $\pm$ 0.9} \\
        & BM25            & 2260.9 $\pm$ 21.7 & 56.8 $\pm$ 1.4 & \underline{56.6 $\pm$ 0.3} & 69.4 $\pm$ 0.2 & \textbf{29.0 $\pm$ 1.1} & 55.5 $\pm$ 1.0 & 53.4 $\pm$ 0.8 \\
        % \cmidrule(lr){2-9}
        % \cmidrule(lr){2-9}
        & ELICIT          & 108.3 $\pm$ 1.4 & 41.6 $\pm$ 0.4 & 46.7 $\pm$ 0.1 & 60.6 $\pm$ 1.4 & 19.1 $\pm$ 1.4 & 49.9 $\pm$ 2.1 & 43.5 $\pm$ 0.8 \\
        & I2CL            & 108.3 $\pm$ 1.4 & 52.4 $\pm$ 4.6 & 48.4 $\pm$ 0.9 & 52.2 $\pm$ 3.1 & 17.9 $\pm$ 2.2 & 45.6 $\pm$ 2.4 & 43.3 $\pm$ 0.7 \\
        & \textbf{ATV}    & 108.3 $\pm$ 1.4 & \textbf{61.0 $\pm$ 5.0} & \textbf{76.1 $\pm$ 1.3} & \textbf{73.0 $\pm$ 1.6} & 25.8 $\pm$ 2.0 & \textbf{74.8 $\pm$ 0.4} & \textbf{62.1 $\pm$ 1.5} \\
        \midrule
        
        \multirow{6}{*}{\textbf{Mistral}}  
        & Zero-shot     & 123.5 $\pm$ 1.7 & 29.6 $\pm$ 1.2 & 26.9 $\pm$ 0.4 & 45.5 $\pm$ 1.3 & 2.8 $\pm$ 0.1 & 36.1 $\pm$ 0.3 & 28.2 $\pm$ 0.5 \\
        & 16-shot       & 2161.3 $\pm$ 0.9 & \underline{55.3 $\pm$ 0.5} & 52.1 $\pm$ 0.5 & \underline{70.8 $\pm$ 0.4} & \underline{23.7 $\pm$ 1.7} & \underline{63.1 $\pm$ 0.6} & 53.0 $\pm$ 0.1 \\
        & BM25          & 2655.2 $\pm$ 27.3 & 55.2 $\pm$ 0.3 & \underline{66.0 $\pm$ 0.5} & 70.2 $\pm$ 1.9 & \textbf{24.1 $\pm$ 0.4} & 62.1 $\pm$ 0.5 & \underline{55.5 $\pm$ 0.4} \\
        % \cmidrule(lr){2-9}
        % \cmidrule(lr){2-9}
        & ELICIT        & 123.5 $\pm$ 1.7 & 41.9 $\pm$ 1.0 & 48.3 $\pm$ 0.3 & 59.4 $\pm$ 0.9 & 20.3 $\pm$ 0.9 & 48.7 $\pm$ 1.8 & 43.7 $\pm$ 0.6 \\
        & I2CL          & 123.5 $\pm$ 1.7 & 48.6 $\pm$ 0.9 & 47.3 $\pm$ 1.5 & 59.6 $\pm$ 0.8 & 17.6 $\pm$ 1.9 & 49.4 $\pm$ 1.0 & 44.5 $\pm$ 0.6 \\
        & \textbf{ATV}  & 123.5 $\pm$ 1.7 & \textbf{60.8 $\pm$ 2.8} & \textbf{69.1 $\pm$ 1.6} & \textbf{71.4 $\pm$ 4.5} & 20.8 $\pm$ 2.4 & \textbf{69.4 $\pm$ 1.9} & \textbf{58.3 $\pm$ 1.3} \\
        \bottomrule
    \end{tabular}
    }
\end{table}
\vspace{-10pt}
\begin{table}[th]
    \centering
    \small

    \caption{
    \textbf{Performance on unseen tasks not included in the ATV training set, evaluated under LLaMA3 and Mistral.} ATV achieves the highest average accuracy across all methods while using significantly fewer tokens than prompt-based and fixed vector approaches, demonstrating strong generalization.  
    All results except I2CL and ATV are from ELICIT~\cite{wang2025elicit}.
    }
    \vspace{0.5em}
    \label{tab:tab_2_unseen}

    \resizebox{\textwidth}{!}{%
        \centering  
        \begin{tabular}{cc|c|ccccc|c}
            \toprule
            \multicolumn{2}{c|}{\textbf{Model}} &\textbf{\# Tokens} & \textbf{GLUE COLA} & \textbf{BBQ Religion} & \textbf{Deepmind} & \textbf{MMLU-Psychology} & \textbf{BBH-five-objects} & \textbf{Avg.} \\ 
            \midrule
            \multirow{5}{*}{\textbf{Llama3}}  
            & Zero-shot     & 103.6 $\pm$ 47.7  & \underline{72.0 $\pm$ 0.7} & 38.6 $\pm$ 1.1 & 17.5 $\pm$ 2.6 & 54.2 $\pm$ 0.3 & 17.1 $\pm$ 0.0 & 39.9 $\pm$ 0.8 \\
            & BM25          & 2502.8 $\pm$ 26.0 & 55.4 $\pm$ 1.0 & \underline{64.6 $\pm$ 1.3} & \textbf{30.7 $\pm$ 1.7} & \textbf{83.0 $\pm$ 0.1} & \underline{48.3 $\pm$ 0.0} & \underline{56.4 $\pm$ 0.4} \\
            & ELICIT        & 103.6 $\pm$ 47.7  & 63.4 $\pm$ 0.9 & 45.0 $\pm$ 0.7 & 23.7 $\pm$ 3.4 & 70.0 $\pm$ 0.6 & 25.7 $\pm$ 0.0 & 45.6 $\pm$ 0.4 \\
            & I2CL          & 103.6 $\pm$ 47.7 & 26.1 $\pm$ 0.6 & 39.4 $\pm$ 3.1 & 23.5 $\pm$ 3.7 & 75.0 $\pm$ 1.0 & 27.3 $\pm$ 2.5 & 38.3 $\pm$ 2.2 \\
            & \textbf{ATV} & 103.6 $\pm$ 47.7 & \textbf{77.6 $\pm$ 2.7} & \textbf{80.8 $\pm$ 2.6} & \underline{26.4 $\pm$ 2.7} & \underline{80.6 $\pm$ 2.3} & \textbf{51.7 $\pm$ 3.1} & \textbf{63.4 $\pm$ 2.5} \\
            \midrule
            \multirow{5}{*}{\textbf{Mistral}}  
            & Zero-shot     & 115.4 $\pm$ 51.0 & 43.3 $\pm$ 1.1 & 35.4 $\pm$ 3.3 & 9.0 $\pm$ 0.4 & 57.9 $\pm$ 0.7 & 7.4 $\pm$ 0.0 & 30.6 $\pm$ 1.0 \\
            & BM25          & 2804.6 $\pm$ 27.6 & 44.4 $\pm$ 2.2 & \underline{70.7 $\pm$ 0.7} & \textbf{26.6 $\pm$ 3.9} & \textbf{78.7 $\pm$ 1.1} & \underline{25.7 $\pm$ 0.0} & \underline{49.2 $\pm$ 0.3} \\
            & ELICIT        & 115.4 $\pm$ 51.0 & 41.7 $\pm$ 0.8 & 42.1 $\pm$ 2.5 & \underline{25.1 $\pm$ 1.2} & 65.6 $\pm$ 0.6 & 15.6 $\pm$ 0.0 & 38.0 $\pm$ 0.6 \\
            & I2CL          & 115.4 $\pm$ 51.0 & \underline{53.3 $\pm$ 1.3} & 48.4 $\pm$ 6.5 & 22.0 $\pm$ 2.6 & \underline{72.6 $\pm$ 0.2} & 22.9 $\pm$ 4.5 & 43.9 $\pm$ 3.0 \\ 
            & \textbf{ATV}  & 115.4 $\pm$ 51.0 & \textbf{79.8 $\pm$ 7.1} & \textbf{81.7 $\pm$ 2.2} & 24.6 $\pm$ 5.3 & 70.7 $\pm$ 1.0 & \textbf{40.3 $\pm$ 3.6} & \textbf{59.4 $\pm$ 2.6} \\
            \bottomrule
        \end{tabular}
    }
\end{table}

We evaluate ATV on 20 in-domain tasks across five categories, with results summarized in Table~\ref{tab:tab_1_main_exp}. ATV consistently achieves the highest average accuracy across all baselines while maintaining strong token efficiency by avoiding additional prompt tokens.
ATV performs particularly well on NLU and Reasoning tasks across both LLaMA3 and Mistral, highlighting the benefit of query-specific task vectors in handling semantic and logical variation. These categories often require nuanced understanding of input structure and are sensitive to prompt formulation, limiting the adaptability of fixed vector approaches. Safety and Knowledge tasks show similar patterns, benefiting from ATV's contextual adaptability.

Notably, BM25 achieves the best performance in the Math category. We attribute this to the pattern-based nature of many math problems, where retrieved demonstrations closely resembling the test query provide a direct advantage. In contrast, ATV’s focus on semantic-level task modeling may limit its effectiveness in tasks that demand precise procedural alignment.
Overall, these results demonstrate the effectiveness of adaptive task representations across diverse language understanding tasks. The findings highlight ATV's particular strength in scenarios requiring contextual adaptation.

\subsection{Generalization to Unseen Tasks}
\label{sec:unseen}

To evaluate the generalization capability of ATV, we assess its performance on a set of unseen tasks held out from training, covering domains such as linguistic acceptability, bias detection, and scientific reasoning.

As shown in Table~\ref{tab:tab_2_unseen}, ATV achieves the highest average accuracy on both LLaMA3 and Mistral, driven by its ability to construct task vectors for novel tasks based on each input even without explicit demonstrations. This result demonstrates the strength of ATV in generalizing beyond in-domain tasks while maintaining strong token efficiency.

\subsection{Ablation Study: Effect of Small Model Capacity}
\label{sec:ablation-small-model}

We study how the capacity of the small language model used to generate task vectors affects ATV's performance.  
All main experiments in this paper use GPT-2 (137M) as the default generator.  
To assess the effect of generator capacity, we experiment with multiple GPT-2 variants while keeping the target model (LLaMA3) fixed, and evaluate on the in-domain benchmark.
As shown in Table~\ref{tab:tab_3_ablation}, larger models such as GPT-2-XL yield slightly better performance, but the gains over GPT-2 are marginal.  
This suggests that even lightweight generators suffice for producing effective task vectors, highlighting the parameter efficiency and practicality of ATV.
\vspace{-10pt}
\begin{table}[ht]
    \centering
    \small

\caption{
    \textbf{Effect of small model capacity on ATV performance.}  
    We evaluate four GPT-2 variants as the small model for generating task vectors, with the LLaMA3 target model fixed.  
    While GPT-2-XL achieves the highest average accuracy, the smallest model (GPT-2, 137M) performs comparably, indicating that lightweight models are sufficient for effective task vector generation and supporting the parameter efficiency of ATV.
}
\label{tab:tab_3_ablation}
\vspace{0.5em}

    \resizebox{\textwidth}{!}{%
        \centering  
        \begin{tabular}{cc|c|ccccc|c}
            \toprule
            \multicolumn{2}{c|}{\textbf{Model}} &\textbf{\# Parameters} & \textbf{NLU} & \textbf{Reasoning} & \textbf{Knowledge} & \textbf{Math} & \textbf{Safety} & \textbf{Avg.} \\ 
            \midrule
            \multirow{4}{*}{\textbf{Llama3}}  
            & GPT-2         & 137M & 61.0 $\pm$ 5.0 & 76.1 $\pm$ 1.3 & 73.0 $\pm$ 1.6 & \textbf{25.8 $\pm$ 2.0} & \textbf{74.8 $\pm$ 0.4} & 62.1 $\pm$ 1.5 \\
            & GPT-2-Medium  & 380M & 61.2 $\pm$ 3.9 & 75.9 $\pm$ 1.4 & 72.0 $\pm$ 2.5 & 24.6 $\pm$ 1.8 & 68.1 $\pm$ 4.1 & 60.4 $\pm$ 1.3 \\
            & GPT-2-Large   & 812M & 62.1 $\pm$ 1.4 & 74.2 $\pm$ 0.6 & 72.4 $\pm$ 2.1 & 25.4 $\pm$ 1.9 & 72.7 $\pm$ 2.8 & 61.4 $\pm$ 0.5  \\
            & GPT-2-XL      & 1.61B & \textbf{63.8 $\pm$ 4.2} & \textbf{76.5 $\pm$ 1.6} & \textbf{73.9 $\pm$ 2.2} & 23.7 $\pm$ 0.8 & 73.1 $\pm$ 4.2 & \textbf{62.2 $\pm$ 0.6} \\
            \bottomrule
        \end{tabular}
    }
\end{table}
\vspace{-10pt}

\subsection{Layer-wise Analysis of Injection Strategies}
\label{sec:layer-wise-analysis}
We conduct a layer-wise analysis to examine how injection depth influences performance for both ATV and ELICIT, revealing distinct patterns in how the two methods interact with different transformer layers. 
While ELICIT requires identifying a single best injection layer per task, we apply uniform injection across different depth ranges, specifically the bottom, middle, and top thirds of the model, to analyze the functional role of each region.

\vspace{-10pt}
\begin{table}[htbp]
    \centering
    \small
    \caption{
    \textbf{Layer-wise performance comparison between ATV and ELICIT on LLaMA3. }
    While ATV shows strong performance when injected into bottom layers, ELICIT performs best when applied only to top layers. 
    This contrast highlights the different functional dependencies of the two methods. Reported differences are measured with respect to the full-layer injection setting.
    }
    \vspace{0.5em}
    \label{tab:tab_4_layer_injection}
    
    \resizebox{0.82\textwidth}{!}{%
        \begin{tabular}{cc|cc|cc}
            \toprule
            \multicolumn{2}{c|}{\textbf{Injected Layer}} 
            & \textbf{Avg. acc (ATV)} & \textbf{Diff. (ATV)} 
            & \textbf{Avg. acc (ELICIT)} & \textbf{Diff. (ELICIT)} \\
            \midrule
            \multirow{4}{*}{\textbf{Llama3}}  
            & All Layers           & \textbf{62.1 $\pm$ 1.5} & -      & 30.9 $\pm$ 0.8 & - \\
            & Bottom $\frac{1}{3}$ & 60.5 $\pm$ 0.7 & \textbf{-1.6}   & 23.5 $\pm$ 0.9 & -17.7 \\
            & Middle $\frac{1}{3}$ & 43.2 $\pm$ 1.0 & -18.9 & 17.8 $\pm$ 0.2 & -18.5 \\
            & Top $\frac{1}{3}$    & 32.6 $\pm$ 0.4 & -29.5 & \textbf{32.8 $\pm$ 0.3} & \textbf{+0.6} \\
            \bottomrule
        \end{tabular}
    }
\end{table}

\begin{wrapfigure}{r}{0.5\textwidth}
    \centering
    \vspace{-14pt}  
    \includegraphics[width=0.45\textwidth]{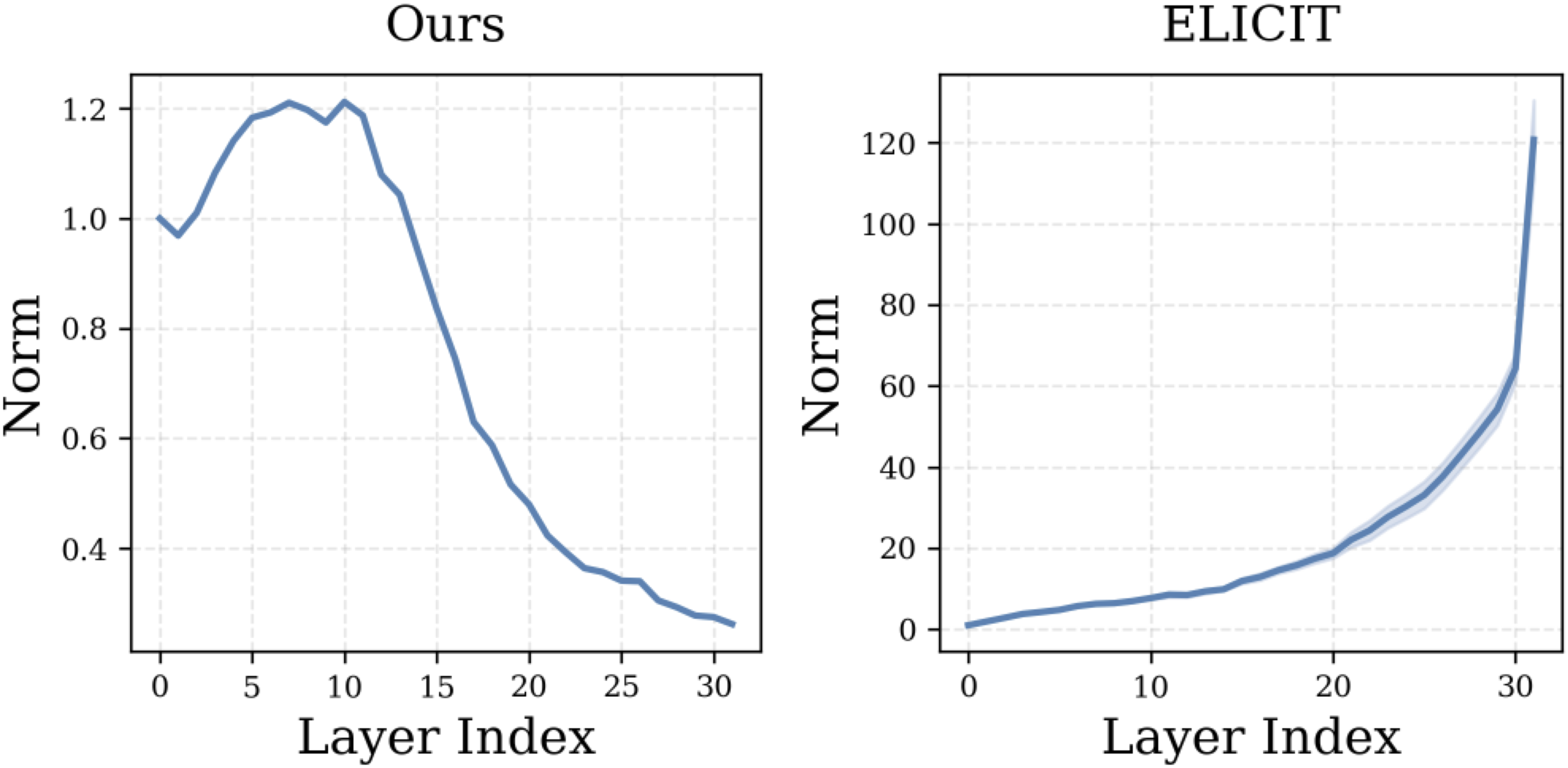}
    % \vspace{-5pt} 
    \caption{\textbf{Layer-wise analysis of vector injection magnitudes.} Left: ATV concentrates vector strength in lower layers, while Right: ELICIT shows a monotonic increase toward top layers. These patterns align with each method's layer-specific performance impact.}
    \label{fig:fig_3_layerwise}
    \vspace{-20pt} 
\end{wrapfigure}

We divide the transformer into bottom, middle, and top thirds, and evaluate each method by restricting injection to a single region.
As shown in Table~\ref{tab:tab_4_layer_injection}, ATV retains strong performance when injected into the bottom layers, exhibiting only marginal degradation relative to the full-layer setting.
For ELICIT, the best performance is observed when injecting into the top layers, slightly surpassing its full-layer setting.
This divergence suggests that ATV benefits from modulating lower-level representations, while ELICIT relies more on upper-layer reasoning.

Figure~\ref{fig:fig_3_layerwise} visualizes the $\ell_2$ norm of the injected vectors across layers. ATV shows higher vector magnitudes in the bottom layers with balanced scales, whereas ELICIT demonstrates a monotonic increase toward the top layers with much larger overall magnitudes.

\begin{wraptable}{r}{0.55\textwidth}
  \vspace{-1.0em}
  \centering
  \caption{
    \textbf{Comparison of LoRA and ATV on LLaMA3.}
    ATV consistently outperforms LoRA on both in-domain and unseen tasks, reflecting its enhanced adaptability. These results are consistent with our theoretical analysis demonstrating the superior expressiveness of ATV.
  }
  \label{tab:tab_5_lora}
  \vspace{0.3em}
  \small
  \renewcommand{\arraystretch}{1.1}
  \begin{tabular}{c|cc|c}
    \toprule
    \textbf{Method} & \textbf{In-domain} & \textbf{Unseen} & \textbf{Avg.} \\
    \midrule
    LoRA & 56.0 $\pm$ 1.4 & 52.0 $\pm$ 3.2 & 54.0 $\pm$ 2.3 \\
    \textbf{ATV} & \textbf{62.1 $\pm$ 1.5} & \textbf{63.4 $\pm$ 2.5} & \textbf{62.8 $\pm$ 2.0} \\
    \bottomrule
  \end{tabular}
  \vspace{-0.7em}
\end{wraptable}

This behavioral difference aligns with the performance trends in Table~\ref{tab:tab_1_main_exp}: ATV achieves consistently strong results across all five task categories, not just in NLU. The effectiveness of early-layer modulation is consistent with prior studies on transformer specialization~\cite{rogers2021primer, tenney2019bert, elhage2021mathematical}, which show that lower layers primarily encode lexical and syntactic features, while upper layers are responsible for semantic reasoning and task-specific abstraction.

In addition, ELICIT exhibits a clear drop in performance when the injection is applied uniformly across all layers without identifying the best-performing layer. This reflects the need for additional computation to determine effective injection locations. By contrast, ATV can be applied across all layers without such tuning, making it more robust and easier to deploy in practice.

\subsection{Empirical Comparison with Low-Rank Adaptation}
\label{sec:compare-lora}

We compare ATV with LoRA~\cite{hu2022lora} on LLaMA3 to validate the theoretical analysis in Section~\ref{sec:theoretical_analysis}. 
As shown in Table~\ref{tab:tab_5_lora}, ATV consistently outperforms LoRA on both in-domain and unseen tasks. These results provide empirical support for our theoretical claim that ATV matches the expressiveness of LoRA while offering better adaptability through query-dependent modulation. In particular, the improvements on unseen tasks highlight ATV’s stronger generalization beyond LoRA’s static update mechanism.

\subsection{Visualizing Task Vector Distributions}
To investigate how task vector distributions from ATV and ELICIT vary with input, we apply t-SNE to visualize the representations generated for different queries.

\begin{figure}[htbp]
    \centering
    \includegraphics[width=0.8\linewidth]{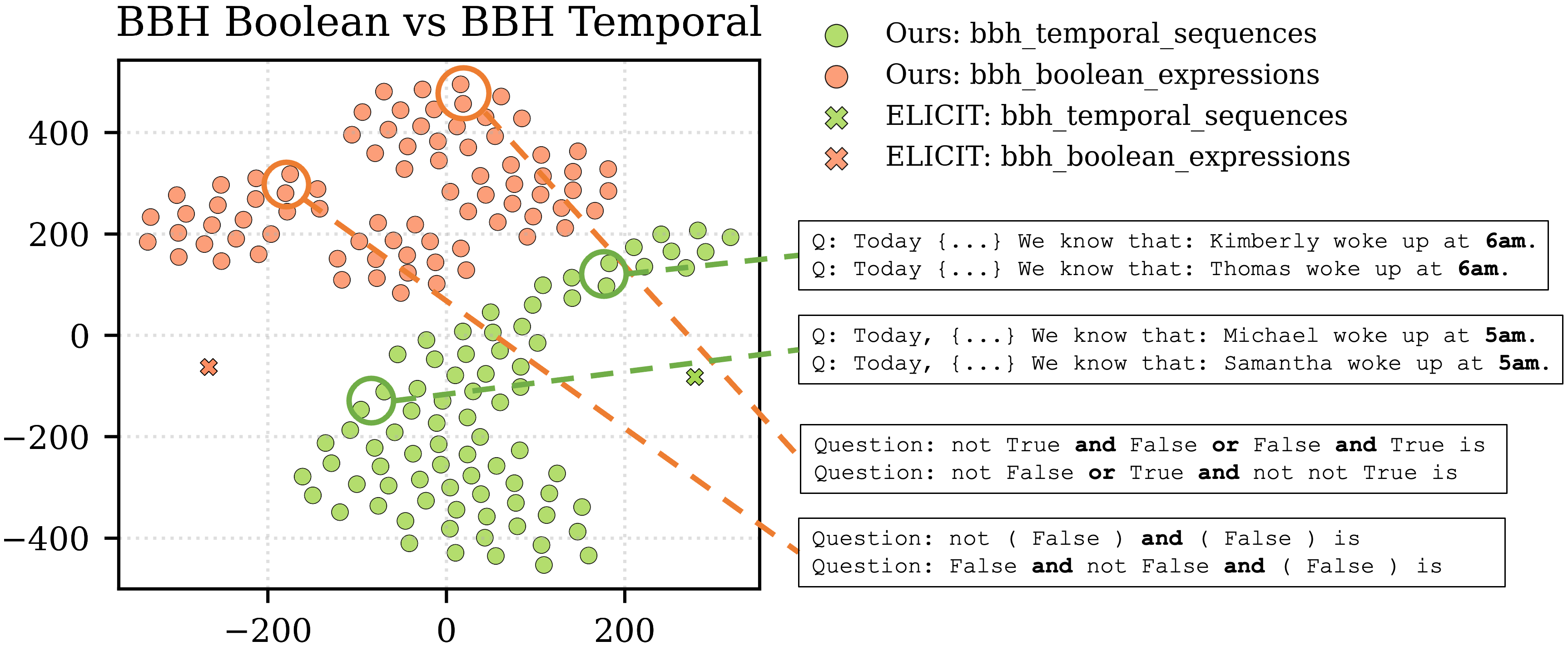}
    \caption{
    \textbf{t-SNE visualization of task vector distributions for two BBH tasks.}
    Each dot represents a query-specific task vector generated by ATV, while crosses denote the fixed task vectors used by ELICIT.
    We observe that vectors from similar queries tend to be grouped together, indicating that ATV adapts its representations based on the input, while ELICIT draws from a fixed demonstration pool and captures less query-level variation as a result.}
    \label{fig:fig_4_tsne_bbh}
\end{figure}

Figure~\ref{fig:fig_4_tsne_bbh} shows the projected vectors for two BBH tasks alongside their ELICIT counterparts.
We observe that, in ATV, vectors from similar queries tend to appear closer together in the embedding space, suggesting that ATV captures input-specific variation within and across tasks.
For ELICIT, while task vectors are retrieved from a fixed pool based on query similarity, the retrieved vectors within each task show limited diversity.
This visualization suggests that ATV produces more input-sensitive task representations, while the retrieval-based vectors used in ELICIT show limited variation across queries within the same task. This representational diversity contributes to ATV's superior performance, consistent with the improvements observed across both in-domain and unseen tasks in Tables \ref{tab:tab_1_main_exp}–\ref{tab:tab_2_unseen}.
\section{Conclusion}
\label{sec:conclusion}

We introduced ATV, a novel framework that addresses fundamental limitations of both ICL and existing task vector approaches. While ICL suffers from computational inefficiency and prior task vector methods rely on fixed representations that fail to adapt to individual inputs, ATV dynamically generates query-specific vectors to modulate LLMs without parameter changes. This approach enables the model to more effectively align with task objectives across diverse input queries.
Empirically, ATV achieves strong performance across both LLaMA3 and Mistral, outperforming prior methods while maintaining high token efficiency. It generalizes well to unseen tasks, demonstrating the effectiveness of query-specific task vector generation without explicit demonstrations. We further support these findings through ablation studies and theoretical analysis, establishing the expressive capacity of ATV and its adaptability across diverse tasks.

\paragraph{Limitation and Impact statement}
Although ATV performs well overall, its effectiveness varies by task, and ATV can sometimes underperform compared to retrieval-based approaches on mathematics tasks where pattern-based matching is more effective than semantic-level task modeling.
Additionally, because ATV depends on the input data, unintended behaviors may occur if the dataset is poorly curated or contains biased examples. Careful dataset understanding and evaluation across diverse input distributions are therefore essential when applying this method in practice.

\smallskip
{
    \small
    \bibliographystyle{IEEEtran}
    \bibliography{ref}

\begin{thebibliography}{10}
\providecommand{\url}[1]{#1}
\csname url@samestyle\endcsname
\providecommand{\newblock}{\relax}
\providecommand{\bibinfo}[2]{#2}
\providecommand{\BIBentrySTDinterwordspacing}{\spaceskip=0pt\relax}
\providecommand{\BIBentryALTinterwordstretchfactor}{4}
\providecommand{\BIBentryALTinterwordspacing}{\spaceskip=\fontdimen2\font plus
\BIBentryALTinterwordstretchfactor\fontdimen3\font minus \fontdimen4\font\relax}
\providecommand{\BIBforeignlanguage}[2]{{%
\expandafter\ifx\csname l@#1\endcsname\relax
\typeout{** WARNING: IEEEtran.bst: No hyphenation pattern has been}%
\typeout{** loaded for the language `#1'. Using the pattern for}%
\typeout{** the default language instead.}%
\else
\language=\csname l@#1\endcsname
\fi
#2}}
\providecommand{\BIBdecl}{\relax}
\BIBdecl

\bibitem{brown2020language}
T.~Brown, B.~Mann, N.~Ryder, M.~Subbiah, J.~D. Kaplan, P.~Dhariwal, A.~Neelakantan, P.~Shyam, G.~Sastry, A.~Askell \emph{et~al.}, ``Language models are few-shot learners,'' \emph{Advances in neural information processing systems}, vol.~33, pp. 1877--1901, 2020.

\bibitem{dong2022survey}
Q.~Dong, L.~Li, D.~Dai, C.~Zheng, J.~Ma, R.~Li, H.~Xia, J.~Xu, Z.~Wu, T.~Liu \emph{et~al.}, ``A survey on in-context learning,'' \emph{arXiv preprint arXiv:2301.00234}, 2022.

\bibitem{liu2021makes}
J.~Liu, D.~Shen, Y.~Zhang, B.~Dolan, L.~Carin, and W.~Chen, ``What makes good in-context examples for gpt-$3 $?'' \emph{arXiv preprint arXiv:2101.06804}, 2021.

\bibitem{peng2024revisiting}
K.~Peng, L.~Ding, Y.~Yuan, X.~Liu, M.~Zhang, Y.~Ouyang, and D.~Tao, ``Revisiting demonstration selection strategies in in-context learning,'' \emph{arXiv preprint arXiv:2401.12087}, 2024.

\bibitem{wang2023large}
X.~Wang, W.~Zhu, and W.~Y. Wang, ``Large language models are implicitly topic models: Explaining and finding good demonstrations for in-context learning,'' \emph{arXiv preprint arXiv:2301.11916}, vol.~1, p.~15, 2023.

\bibitem{li2024long}
T.~Li, G.~Zhang, Q.~D. Do, X.~Yue, and W.~Chen, ``Long-context llms struggle with long in-context learning,'' \emph{arXiv preprint arXiv:2404.02060}, 2024.

\bibitem{kuratov2024babilong}
Y.~Kuratov, A.~Bulatov, P.~Anokhin, I.~Rodkin, D.~Sorokin, A.~Sorokin, and M.~Burtsev, ``Babilong: Testing the limits of llms with long context reasoning-in-a-haystack,'' \emph{Advances in Neural Information Processing Systems}, vol.~37, pp. 106\,519--106\,554, 2024.

\bibitem{hendel-etal-2023-context}
R.~Hendel, M.~Geva, and A.~Globerson, ``In-context learning creates task vectors,'' in \emph{Findings of the Association for Computational Linguistics: EMNLP 2023}, H.~Bouamor, J.~Pino, and K.~Bali, Eds., Dec. 2023, pp. 9318--9333.

\bibitem{ilharco2023editing}
G.~Ilharco, M.~T. Ribeiro, M.~Wortsman, L.~Schmidt, H.~Hajishirzi, and A.~Farhadi, ``Editing models with task arithmetic,'' in \emph{The Eleventh International Conference on Learning Representations}, 2023.

\bibitem{liu2024context}
S.~Liu, H.~Ye, L.~Xing, and J.~Zou, ``In-context vectors: making in context learning more effective and controllable through latent space steering,'' in \emph{Proceedings of the 41st International Conference on Machine Learning}, 2024, pp. 32\,287--32\,307.

\bibitem{li2025implicit}
Z.~Li, Z.~Xu, L.~Han, Y.~Gao, S.~Wen, D.~Liu, H.~Wang, and D.~N. Metaxas, ``Implicit in-context learning,'' in \emph{The Thirteenth International Conference on Learning Representations}, 2025.

\bibitem{wang2025elicit}
F.~Wang, J.~Yan, Y.~Zhang, and T.~Lin, ``{ELICIT}: {LLM} augmentation via external in-context capability,'' in \emph{The Thirteenth International Conference on Learning Representations}, 2025.

\bibitem{yang2025task}
L.~Yang, Z.~Lin, K.~Lee, D.~Papailiopoulos, and R.~Nowak, ``Task vectors in in-context learning: Emergence, formation, and benefit,'' \emph{arXiv preprint arXiv:2501.09240}, 2025.

\bibitem{huang2024multimodal}
B.~Huang, C.~Mitra, L.~Karlinsky, A.~Arbelle, T.~Darrell, and R.~Herzig, ``Multimodal task vectors enable many-shot multimodal in-context learning,'' \emph{Advances in Neural Information Processing Systems}, vol.~37, pp. 22\,124--22\,153, 2024.

\bibitem{zhao2021calibrate}
Z.~Zhao, E.~Wallace, S.~Feng, D.~Klein, and S.~Singh, ``Calibrate before use: Improving few-shot performance of language models,'' in \emph{International conference on machine learning}.\hskip 1em plus 0.5em minus 0.4em\relax PMLR, 2021, pp. 12\,697--12\,706.

\bibitem{lu2021fantastically}
Y.~Lu, M.~Bartolo, A.~Moore, S.~Riedel, and P.~Stenetorp, ``Fantastically ordered prompts and where to find them: Overcoming few-shot prompt order sensitivity,'' \emph{arXiv preprint arXiv:2104.08786}, 2021.

\bibitem{zhang2024batch}
K.~Zhang, A.~Lv, Y.~Chen, H.~Ha, T.~Xu, and R.~Yan, ``Batch-icl: Effective, efficient, and order-agnostic in-context learning,'' \emph{arXiv preprint arXiv:2401.06469}, 2024.

\bibitem{li2025task}
H.~Li, Y.~Zhang, S.~Zhang, M.~Wang, S.~Liu, and P.-Y. Chen, ``When is task vector provably effective for model editing? a generalization analysis of nonlinear transformers,'' \emph{arXiv preprint arXiv:2504.10957}, 2025.

\bibitem{wei2022chain}
J.~Wei, X.~Wang, D.~Schuurmans, M.~Bosma, F.~Xia, E.~Chi, Q.~V. Le, D.~Zhou \emph{et~al.}, ``Chain-of-thought prompting elicits reasoning in large language models,'' \emph{Advances in neural information processing systems}, vol.~35, pp. 24\,824--24\,837, 2022.

\bibitem{kojima2022large}
T.~Kojima, S.~S. Gu, M.~Reid, Y.~Matsuo, and Y.~Iwasawa, ``Large language models are zero-shot reasoners,'' \emph{Advances in neural information processing systems}, vol.~35, pp. 22\,199--22\,213, 2022.

\bibitem{zhou2022least}
D.~Zhou, N.~Sch{\"a}rli, L.~Hou, J.~Wei, N.~Scales, X.~Wang, D.~Schuurmans, C.~Cui, O.~Bousquet, Q.~Le \emph{et~al.}, ``Least-to-most prompting enables complex reasoning in large language models,'' \emph{arXiv preprint arXiv:2205.10625}, 2022.

\bibitem{min2022rethinking}
S.~Min, X.~Lyu, A.~Holtzman, M.~Artetxe, M.~Lewis, H.~Hajishirzi, and L.~Zettlemoyer, ``Rethinking the role of demonstrations: What makes in-context learning work?'' in \emph{Proceedings of the 2022 Conference on Empirical Methods in Natural Language Processing}, 2022, pp. 11\,048--11\,064.

\bibitem{rubin2022learning}
O.~Rubin, J.~Herzig, and J.~Berant, ``Learning to retrieve prompts for in-context learning,'' in \emph{Proceedings of the 2022 Conference of the North American Chapter of the Association for Computational Linguistics: Human Language Technologies}, 2022, pp. 2655--2671.

\bibitem{vaswani2017attention}
A.~Vaswani, N.~Shazeer, N.~Parmar, J.~Uszkoreit, L.~Jones, A.~N. Gomez, {\L}.~Kaiser, and I.~Polosukhin, ``Attention is all you need,'' \emph{Advances in neural information processing systems}, vol.~30, 2017.

\bibitem{radford2019language}
A.~Radford, J.~Wu, R.~Child, D.~Luan, D.~Amodei, I.~Sutskever \emph{et~al.}, ``Language models are unsupervised multitask learners,'' \emph{OpenAI blog}, vol.~1, no.~8, p.~9, 2019.

\bibitem{hu2022lora}
E.~J. Hu, Y.~Shen, P.~Wallis, Z.~Allen-Zhu, Y.~Li, S.~Wang, L.~Wang, W.~Chen \emph{et~al.}, ``Lora: Low-rank adaptation of large language models.'' \emph{ICLR}, vol.~1, no.~2, p.~3, 2022.

\bibitem{li2021prefix}
X.~L. Li and P.~Liang, ``Prefix-tuning: Optimizing continuous prompts for generation,'' \emph{arXiv preprint arXiv:2101.00190}, 2021.

\bibitem{dai2022can}
D.~Dai, Y.~Sun, L.~Dong, Y.~Hao, S.~Ma, Z.~Sui, and F.~Wei, ``Why can gpt learn in-context? language models implicitly perform gradient descent as meta-optimizers,'' \emph{arXiv preprint arXiv:2212.10559}, 2022.

\bibitem{he2022towards}
J.~He, C.~Zhou, X.~Ma, T.~Berg-Kirkpatrick, and G.~Neubig, ``Towards a unified view of parameter-efficient transfer learning,'' in \emph{International Conference on Learning Representations}, 2022.

\bibitem{grattafiori2024llama}
A.~Grattafiori, A.~Dubey, A.~Jauhri, A.~Pandey, A.~Kadian, A.~Al-Dahle, A.~Letman, A.~Mathur, A.~Schelten, A.~Vaughan \emph{et~al.}, ``The llama 3 herd of models,'' \emph{arXiv preprint arXiv:2407.21783}, 2024.

\bibitem{jiang2023mistral7b}
\BIBentryALTinterwordspacing
A.~Q. Jiang, A.~Sablayrolles, A.~Mensch, C.~Bamford, D.~S. Chaplot, D.~de~las Casas, F.~Bressand, G.~Lengyel, G.~Lample, L.~Saulnier, L.~R. Lavaud, M.-A. Lachaux, P.~Stock, T.~L. Scao, T.~Lavril, T.~Wang, T.~Lacroix, and W.~E. Sayed, ``Mistral 7b,'' 2023. [Online]. Available: \url{https://arxiv.org/abs/2310.06825}
\BIBentrySTDinterwordspacing

\bibitem{robertson2009probabilistic}
S.~Robertson, H.~Zaragoza \emph{et~al.}, ``The probabilistic relevance framework: Bm25 and beyond,'' \emph{Foundations and Trends{\textregistered} in Information Retrieval}, vol.~3, no.~4, pp. 333--389, 2009.

\bibitem{rogers2021primer}
A.~Rogers, O.~Kovaleva, and A.~Rumshisky, ``A primer in bertology: What we know about how bert works,'' \emph{Transactions of the association for computational linguistics}, vol.~8, pp. 842--866, 2021.

\bibitem{tenney2019bert}
I.~Tenney, D.~Das, and E.~Pavlick, ``Bert rediscovers the classical nlp pipeline,'' \emph{arXiv preprint arXiv:1905.05950}, 2019.

\bibitem{elhage2021mathematical}
N.~Elhage, N.~Nanda, C.~Olsson, T.~Henighan, N.~Joseph, B.~Mann, A.~Askell, Y.~Bai, A.~Chen, T.~Conerly \emph{et~al.}, ``A mathematical framework for transformer circuits,'' \emph{Transformer Circuits Thread}, vol.~1, no.~1, p.~12, 2021.

\bibitem{talmor2018commonsenseqa}
A.~Talmor, J.~Herzig, N.~Lourie, and J.~Berant, ``Commonsenseqa: A question answering challenge targeting commonsense knowledge,'' \emph{arXiv preprint arXiv:1811.00937}, 2018.

\bibitem{mihaylov2018can}
T.~Mihaylov, P.~Clark, T.~Khot, and A.~Sabharwal, ``Can a suit of armor conduct electricity? a new dataset for open book question answering,'' \emph{arXiv preprint arXiv:1809.02789}, 2018.

\bibitem{zellers2019hellaswag}
R.~Zellers, A.~Holtzman, Y.~Bisk, A.~Farhadi, and Y.~Choi, ``Hellaswag: Can a machine really finish your sentence?'' \emph{arXiv preprint arXiv:1905.07830}, 2019.

\bibitem{clark2019boolq}
C.~Clark, K.~Lee, M.-W. Chang, T.~Kwiatkowski, M.~Collins, and K.~Toutanova, ``Boolq: Exploring the surprising difficulty of natural yes/no questions,'' \emph{arXiv preprint arXiv:1905.10044}, 2019.

\bibitem{suzgun2022challenging}
M.~Suzgun, N.~Scales, N.~Sch{\"a}rli, S.~Gehrmann, Y.~Tay, H.~W. Chung, A.~Chowdhery, Q.~V. Le, E.~H. Chi, D.~Zhou \emph{et~al.}, ``Challenging big-bench tasks and whether chain-of-thought can solve them,'' \emph{arXiv preprint arXiv:2210.09261}, 2022.

\bibitem{clark2018think}
P.~Clark, I.~Cowhey, O.~Etzioni, T.~Khot, A.~Sabharwal, C.~Schoenick, and O.~Tafjord, ``Think you have solved question answering? try arc, the ai2 reasoning challenge,'' \emph{arXiv preprint arXiv:1803.05457}, 2018.

\bibitem{amini2019mathqa}
A.~Amini, S.~Gabriel, P.~Lin, R.~Koncel-Kedziorski, Y.~Choi, and H.~Hajishirzi, ``Mathqa: Towards interpretable math word problem solving with operation-based formalisms,'' \emph{arXiv preprint arXiv:1905.13319}, 2019.

\bibitem{wang2024mmlu}
Y.~Wang, X.~Ma, G.~Zhang, Y.~Ni, A.~Chandra, S.~Guo, W.~Ren, A.~Arulraj, X.~He, Z.~Jiang \emph{et~al.}, ``Mmlu-pro: A more robust and challenging multi-task language understanding benchmark,'' in \emph{The Thirty-eight Conference on Neural Information Processing Systems Datasets and Benchmarks Track}, 2024.

\bibitem{nangia2020crows}
N.~Nangia, C.~Vania, R.~Bhalerao, and S.~R. Bowman, ``Crows-pairs: A challenge dataset for measuring social biases in masked language models,'' \emph{arXiv preprint arXiv:2010.00133}, 2020.

\bibitem{parrish2021bbq}
A.~Parrish, A.~Chen, N.~Nangia, V.~Padmakumar, J.~Phang, J.~Thompson, P.~M. Htut, and S.~R. Bowman, ``Bbq: A hand-built bias benchmark for question answering,'' \emph{arXiv preprint arXiv:2110.08193}, 2021.

\bibitem{merity2016pointer}
S.~Merity, C.~Xiong, J.~Bradbury, and R.~Socher, ``Pointer sentinel mixture models,'' \emph{arXiv preprint arXiv:1609.07843}, 2016.

\bibitem{wang2018glue}
A.~Wang, A.~Singh, J.~Michael, F.~Hill, O.~Levy, and S.~R. Bowman, ``Glue: A multi-task benchmark and analysis platform for natural language understanding,'' \emph{arXiv preprint arXiv:1804.07461}, 2018.

\bibitem{wang2019superglue}
A.~Wang, Y.~Pruksachatkun, N.~Nangia, A.~Singh, J.~Michael, F.~Hill, O.~Levy, and S.~Bowman, ``Superglue: A stickier benchmark for general-purpose language understanding systems,'' \emph{Advances in neural information processing systems}, vol.~32, 2019.

\bibitem{saxton2019analysing}
D.~Saxton, E.~Grefenstette, F.~Hill, and P.~Kohli, ``Analysing mathematical reasoning abilities of neural models,'' \emph{arXiv preprint arXiv:1904.01557}, 2019.

\bibitem{lintang_sutawika_2023_10256836}
\BIBentryALTinterwordspacing
L.~Sutawika, L.~Gao, H.~Schoelkopf, S.~Biderman, J.~Tow, B.~Abbasi, ben fattori, C.~Lovering, farzanehnakhaee70, J.~Phang, A.~Thite, Fazz, Aflah, N.~Muennighoff, T.~Wang, sdtblck, nopperl, gakada, tttyuntian, researcher2, Chris, J.~Etxaniz, Z.~Kasner, Khalid, J.~Hsu, AndyZwei, P.~S. Ammanamanchi, D.~Groeneveld, E.~Smith, and E.~Tang, ``Eleutherai/lm-evaluation-harness: Major refactor,'' Dec. 2023. [Online]. Available: \url{https://doi.org/10.5281/zenodo.10256836}
\BIBentrySTDinterwordspacing

\end{thebibliography}
}

%%%%%%%%%%%%%%%%%%%%%%%%%%%%%%%%%%%%%%%%%%%%%%%%%%%%%%%%%%%%

\newpage
\clearpage

\appendix

\section{Proofs for Theoretical Results}
\label{app:proof}

\subsection{Proof of Theorem 1}
\label{app:proof_thrm1}

\subsubsection{Preliminaries}
\textit{Consider a single transformer layer $\ell$ with hidden state dimension $d_\ell$.}
Let $h^\ell \in \mathbb{R}^{d_\ell}$ denote the hidden state of the last token at that layer.

\paragraph{ATV (Adaptive Task Vector) update}
ATV injects an additive low-rank vector
\[
    \tilde{h}^\ell = h^\ell + \lambda\, v_{\text{small}} A_\ell,
\]
where
\begin{itemize}
    \item $h^\ell \in \mathbb{R}^{d_\ell}$: last-token hidden state in layer $\ell$.
    \item $v_{\text{small}} \in \mathbb{R}^{d_s}$ with $d_s \ll d_\ell$: query-specific vector from the \( \mathcal{M}_{\text{small}} \).
    \item $A_\ell \in \mathbb{R}^{d_s \times d_\ell}$: the $\ell$-th $d_s \times d_\ell$ block of the linear expansion $f_\theta(v_{\text{small}}) = v_{\text{ATV}} \in \mathbb{R}^{L \times d_\ell}$; thus, $v_{\text{small}} A_\ell$ is the $\ell$-th row of $v_{\text{ATV}}$.
    \item $\lambda \in \mathbb{R}$: global scaling constant.
\end{itemize}

\paragraph{LoRA (low-rank adaptation) update}
LoRA applies a rank-$r$ additive update to the projection. Its effect on the hidden state is:
\[
    \hat{h}^\ell = h^\ell + s\, x^\top W_{\text{down}} W_{\text{up}},
\]
where
\begin{itemize}
    \item $x^\top \in \mathbb{R}^{1 \times d_\ell}$: the current-token activation entering the augmented projection; for a query / value projection, this equals $(h^\ell)^\top$.
    \item $W_{\text{down}} \in \mathbb{R}^{d_\ell \times r}$ and $W_{\text{up}} \in \mathbb{R}^{r \times d_\ell}$: LoRA projection matrices.
    \item $s \in \mathbb{R}$: global scaling constant.
\end{itemize}

\paragraph{Assumption (A1) — matched rank budgets}
We henceforth set the LoRA bottleneck rank equal to the ATV bottleneck size:
\[
    r = d_s.
\]

\subsubsection{Theorem}
\paragraph{Theorem (ATV--LoRA equivalence under equal rank)}
Under Assumption A1:
\begin{enumerate}
    \item \textbf{ATV $\Rightarrow$ LoRA (simulation).} \\
    For every pair $(h^\ell, v_{\text{small}})$ there exist static LoRA factors $(W_{\text{down}}, W_{\text{up}})$ and a scale $s$, all independent of the runtime query, such that
    \[
        \tilde{h}^\ell = \hat{h}^\ell \quad \text{for all inputs}.
    \]
    
    \item \textbf{LoRA $\Rightarrow$ ATV (simulation).} \\
    Conversely, any LoRA update with rank $r = d_s$ can be expressed in ATV form by an appropriate choice of $(\lambda, v_{\text{small}}, A_\ell)$.
\end{enumerate}

Hence, when the rank budgets are matched, ATV and LoRA realize the same class of low-rank additive perturbations to the frozen model; they are expressively equivalent.

\subsubsection{Proof}
\label{app:proof_thrm1_proof}
\textit{Throughout the proof we fix the layer index $\ell$ and omit it from superscripts whenever no ambiguity arises.}

\paragraph{Step 1 \quad Factorize the ATV increment}
The additive term introduced by ATV is
\[
\Delta h_{\text{ATV}} = \lambda\, v_{\text{small}} A_\ell
= \underbrace{(\lambda v_{\text{small}})}_{ 1 \times d_s}
\underbrace{A_\ell}_{ d_s \times d_\ell}.
\]
This is an outer product of a $1 \times d_s$ row vector and a $d_s \times d_\ell$ matrix, so its matrix rank satisfies
\[
\operatorname{rank}(\Delta h_{\text{ATV}}) \leq d_s.
\]
Hence, ATV always adds a vector lying in a rank-$d_s$ subspace of $\mathbb{R}^{d_\ell}$.

\paragraph{Step 2 \quad Factorize the LoRA increment}
LoRA's contribution can be written in vector form as
\[
\Delta h_{\text{LoRA}} = s\, x^\top W_{\text{down}} W_{\text{up}}
= \underbrace{(s\, x^\top W_{\text{down}})}_{1 \times r}
\underbrace{W_{\text{up}}}_{r \times d_\ell},
\]
which is likewise an outer product—now of shapes $1 \times r$ and $r \times d_\ell$. Consequently,
\[
\operatorname{rank}(\Delta h_{\text{LoRA}}) \leq r.
\]
Under Assumption A1 ($r = d_s$), the rank upper bounds obtained in Steps 1 and 2 are identical,  
establishing that the two increments live in subspaces of equal maximal rank.

\paragraph{Step 3 \quad Rewrite ATV as a LoRA update (ATV $\Rightarrow$ LoRA)}

Let
\[
W_{\text{up}} := A_\ell \quad \text{and} \quad W_{\text{down}} := M \in \mathbb{R}^{d_\ell \times d_s}.
\]
We choose $M$ so that the following linear constraint holds for the last token:
\begin{equation}
    x^\top M = \lambda\, v \in \mathbb{R}^{1 \times d_s}.
\end{equation}

\textbf{Existence of a solution.}
The row vector $x^\top$ has $d_\ell$ free coordinates, whereas the right-hand side specifies only $d_s$ values with $d_s \ll d_\ell$; consequently, the under-determined system (1) always admits a solution provided $x^\top \neq 0$. A canonical choice is
\[
M := x^{+\top} (\lambda v),
\]
where $x^{+\top}$ denotes the Moore–Penrose pseudoinverse of $x^\top$ and satisfies $x^\top x^{+\top} = 1$.

\textbf{Resulting LoRA increment.}
With these matrices,
\[
\Delta h_{\text{LoRA}} = s\, x^\top W_{\text{down}} W_{\text{up}} = s\, (x^\top M)\, A_\ell = s\, (\lambda v)\, A_\ell.
\]
Selecting the scale $s = 1$ yields
\[
\Delta h_{\text{LoRA}} = \lambda v A_\ell = \Delta h_{\text{ATV}},
\]
and hence the LoRA-modified hidden state satisfies $\tilde{h}^\ell = \hat{h}^\ell$.

This completes the direction ``ATV implies LoRA'' under the rank-matching assumption $r = d_s$.

\paragraph{Step 4 \quad Rewrite LoRA as an ATV update (LoRA $\Rightarrow$ ATV)}

Take fixed LoRA factors $W_{\text{down}} \in \mathbb{R}^{d_\ell \times r}$, $W_{\text{up}} \in \mathbb{R}^{r \times d_\ell}$ with $r = d_s$.

For the last token we observe the row vector
\[
x^\top W_{\text{down}} \in \mathbb{R}^{1 \times d_s}.
\]

\textbf{Row-vector SVD.}  
Compute a thin singular-value decomposition
\[
x^\top W_{\text{down}} = u \Sigma v^\top,
\]
where
\begin{itemize}
    \item $u \in \mathbb{R}^{1 \times d_s}$ is orthonormal,
    \item $\Sigma \in \mathbb{R}^{d_s \times d_s}$ is diagonal (rank $\leq 1$),
    \item $v^\top \in \mathbb{R}^{d_s \times d_s}$ is orthonormal.
\end{itemize}

\textbf{Define ATV parameters}  
\[
\lambda := \|u \Sigma\|_2, \quad
v := \frac{u \Sigma}{\|u \Sigma\|_2} \in \mathbb{R}^{d_s}, \quad
A_\ell := v^\top W_{\text{up}} \in \mathbb{R}^{d_s \times d_\ell}.
\]

The vector $v$ has unit norm by construction, and $A_\ell$ keeps the required shape.

\textbf{ATV increment equals LoRA increment}  
\[
\lambda v A_\ell
= \left( \|u \Sigma\|_2 \right) \frac{u \Sigma}{\|u \Sigma\|_2} v^\top W_{\text{up}}
= (u \Sigma)\, v^\top W_{\text{up}}
= x^\top W_{\text{down}} W_{\text{up}}
= \Delta h_{\text{LoRA}}.
\]

Hence $\tilde{h}^\ell = \hat{h}^\ell$; the LoRA update is reproduced exactly by ATV.

\paragraph{Conclusion of the proof}
Both ATV and LoRA ultimately realize the same operation class
\[
    h \longmapsto h + (\text{rank} \leq r)\ \text{outer product},
\]
and, under the matched rank budget $r = d_s$, have identical functional capacity on a frozen backbone.  
This completes the proof of the theorem.

\subsubsection{Discussion of Assumptions}

\begin{itemize}
    \item \textbf{Matched rank budgets ($r = d_s$).} \\
    The constructive proof fixes the LoRA rank to equal the ATV bottleneck size so that both methods have the same number of degrees of freedom.

    \textit{If $d_s < r$}, LoRA has $r - d_s$ surplus channels; turning those channels off yields an exact simulation of ATV, so LoRA is (weakly) more expressive.

    \textit{If $d_s > r$}, ATV can move in directions that LoRA cannot represent. Projecting the ATV increment onto a rank-$r$ subspace gives the closest LoRA-matchable update, so equivalence holds only ``up to rank-$r$ projection.''

    \item \textbf{Non-zero input row vector $x^\top$.} \\
    The linear system $x^\top M = \lambda v$ in Step 3 requires $x^\top \neq 0$ to admit a solution via the Moore–Penrose pseudoinverse.

    In a transformer, $x^\top$ is simply the hidden state of the token entering the query/value projection and is never identically zero after normal training; therefore the assumption is benign in practice.
\end{itemize}

\subsubsection{Operational Differences Between ATV and LoRA}

\begin{table}[h]
\centering
\caption{Implementation-level differences between ATV and LoRA.}
\vspace{0.5em}
\resizebox{\textwidth}{!}{%
\begin{tabular}{@{}p{2.6cm} p{5.5cm} p{5.5cm}@{}}
\toprule
\textbf{Aspect} & \textbf{ATV} & \textbf{LoRA} \\
\midrule
\textit{Insertion point} &
Adds a vector $\lambda v A_\ell$ \textit{after} the hidden state is computed &
Adds a low-rank matrix $W_{\text{down}} W_{\text{up}}$ \textit{inside} the projection weight \\
\addlinespace[0.5ex]
\textit{Learned parameters} &
Auxiliary small model \( \mathcal{M}_{\text{small}} \) $+$ expansion blocks $A_\ell$ of the linear expansion $f_\theta()$ &
Two fixed factor matrices $W_{\text{down}}, W_{\text{up}}$ \\
\bottomrule
\end{tabular}
}
\end{table}

Thus, even under equal rank budgets, the two methods differ operationally:
ATV perturbs activations directly in the hidden state space, whereas LoRA perturbs projection weights via a static low-rank matrix. Nevertheless, as shown in the theorem, these implementation choices realize the same class of rank-$r$ additive perturbations on a frozen backbone, yielding identical expressive power when $r = d_s$.

\subsection{Proof of Theorem 2}
\label{app:proof_thrm2}

\subsubsection{Preliminaries}
\textbf{Linear approximation:} $\text{Attn}(Q,K,V) \approx QK^\top V$ ~\cite{dai2022can}

\paragraph{Attention ouptputs from Prefix-Tuning}
\[
\text{Attn}_{\text{prefix}} = \text{Attn}(x W_q,[P_k; CW_k], [P_v; CW_v])
\]
\begin{itemize}
  \item $x=[x_1, x_2, ...,x_T] \in \mathbb{R}^{T \times d_l}$
  \item $C \in \mathbb{R}^{m \times d_l}$: \\
  context sequence of length $m$ with $d_l$ dimension($l$-th layer's dimension)
  \item $P_k,P_v \in \mathbb{R}^{p \times d_l} $: $p$ tunable prefix vectors to the keys and values
\end{itemize}

\paragraph{Attention ouptputs from Prefix-Tuning}
\[
\text{Attn}_{\text{ATV}} = \text{Attn}((x+e_{T} \cdot ({v_{ATV}^l})^\top )W_q,(C+e_{m} \cdot ({v_{ATV}^l})^\top)W_k,(C+e_{m} \cdot ({v_{ATV}^l})^\top)W_v)
\]
\begin{itemize}
  \item $v_{ATV}^l \in \mathbb{R}^{d_l}$
  \item $e_{m} = [0, ..., 0, 1] \in \mathbb{R}^{m \times 1}$
  \item $(e_{m} \cdot ({v_{ATV}^l})^\top)W_k = P^{\prime}_k$ 
\end{itemize}

\subsubsection{Theorem}
\paragraph{Theorem (ATV is more expressive than Prefix-Tuning)}
The representational space $\mathcal{F}$ of $\text{Attn}_{\text{ATV}}$ includes that of $\text{Attn}_{\text{prefix}}$:
\[
\mathcal{F}(\text{Attn}_{\text{prefix}}) \subseteq \mathcal{F}(\text{Attn}_{\text{ATV}})
\]

\subsubsection{Process of Derivation}
\paragraph{Linear approximation of Prefix-Tuning}
\begin{align*}
\text{Attn}_{\text{prefix}}
&=\text{Attn}(x W_q,concat(P_k,CW_k), concat(P_v,CW_v)) \\
&=\text{softmax}(xW_q ([P_k;CW_k])^\top) [P_v;CW_v]  \\
&\approx xW_q(P_k;CW_k])^\top ([P_v;CW_v]) \space ( \because \text{Attn}(Q,K,V) \approx QK^\top V \text) \\
&= xW_q(P_k)^\top P_v + xW_q (CW_k)^\top CW_v
\end{align*}

\paragraph{Linear approximation of ATV}
\begin{align*}
\text{Attn}_{\text{ATV}}=
&\text{Attn}\left((x + e_{T} \cdot ({v_{ATV}^l})^\top) W_q,\ (C + e_{m} \cdot ({v_{ATV}^l})^\top) W_k,\ (C + e_{m} \cdot ({v_{ATV}^l})^\top) W_v\right) \\
&\approx \Big[\, x W_q (C W_k)^\top 
+ x W_q (e_{m} \cdot ({v_{ATV}^l})^\top)^\top \\
&\quad + e_{T} \cdot ({v_{ATV}^l})^\top W_q (C W_k)^\top 
+ e_{T} \cdot ({v_{ATV}^l})^\top W_q (e_{m} \cdot ({v_{ATV}^l})^\top W_k)^\top\, \Big] \\
&\quad \cdot \left(C W_v + e_{m} \cdot ({v_{ATV}^l})^\top W_v\right)
\end{align*}

\noindent\textbf{Let } $(e_{m} \cdot ({v_{ATV}^l})^\top) W_k = P_k'$, $(e_{m} \cdot ({v_{ATV}^l})^\top) W_v = P_v'$, $(e_{T} \cdot ({v_{ATV}^l})^\top) W_q = P^{\prime}_q$

\begin{align*}
\Rightarrow\quad & \text{Attn}_{\text{ATV}} \approx 
x W_q (C W_k)^\top C W_v
+ x W_q (P_k')^\top P_v' \space (\text{Similar to }\text{Attn}_\text{Prefix}) \tag{$T_1 +T_2$}\\
& \quad \quad \quad \quad + x W_q ({P^{\prime}_k})^\top C W_v \tag{$T_3$} \\
& \quad \quad \quad \quad + x W_q(C W_k)^\top P^{\prime}_v \tag{$T_4$}\\
&\quad \quad \quad \quad+ P^{\prime}_q (C W_k)^\top C W_v \tag{$T_5$}\\
&\quad \quad \quad \quad+ P^{\prime}_q({P^{\prime}_k})^\top C W_v \tag{$T_6$}\\
&\quad \quad \quad \quad+ P^{\prime}_q(C W_k)^\top P^{\prime}_v \tag{$T_7$}\\
&\quad \quad \quad \quad+ P^{\prime}_q{P^{\prime}_k}^\top P^{\prime}_v \tag{$T_8$}
\end{align*} 

\subsubsection{Analysis of Each Term in $\text{Attn}_{\text{ATV}}$}
For each term we report (i) the intuition behind the interaction,
and (ii) how it extends or subsumes the behavior attainable with classic
Prefix-Tuning (PT), treating the ATV-generated vectors
$P^{\prime}_{k}, P^{\prime}_{v}, P^{\prime}_{q}$ as soft-prefix counterparts to PT’s fixed prefixes
(see Table~\ref{tab:atv_terms}).

\begin{table}[ht]
\centering
\caption{Qualitative roles of ATV attention terms and their relation to Prefix-Tuning (PT).} 
\label{tab:atv_terms}
\vspace{0.5em}

\renewcommand{\arraystretch}{1.4}
\begin{tabular}{@{}>{\bfseries}c p{5cm} p{8cm}@{}}
\toprule
\textbf{Term} & \textbf{Qualitative role} & \textbf{Relation to Prefix-Tuning (PT) / Added expressivity} \\
\midrule

$T_1$ & \textbf{Base attention} of the frozen model & \textbf{Representable in PT (identical).} No additional expressivity; both methods preserve this term unchanged. \\

$T_2$ & \textbf{Prefix keys \& values only.} Query attends only to prefix key/value & \textbf{Representable in PT (exact match).} This is the sole extra path PT can realize; ATV contains it and can therefore emulate PT exactly. \\

$T_3$ & \textbf{Prefix key → content values.} A soft prefix key reshapes the attention weights, but the actual information still comes from the content values. & \textbf{Not representable in PT.} PT would need a separate learned key for every content token, whereas ATV achieves the same effect with a single soft key—thus widening the attention design space. \\

$T_4$ & \textbf{Content keys → prefix value.} Normal keys set the weights, but an extra value generated by the adapter is injected at the output stage. & \textbf{Not representable in PT.} PT lacks a mechanism to inject new information exclusively at the value stage for existing keys; ATV can graft auxiliary content into any token’s output. \\

$T_5$ & \textbf{Prefix query.} The query itself is shifted in a new direction while still using ordinary keys and values. & \textbf{Not representable in PT.} Because PT keeps $W_q$ frozen, it cannot alter queries. ATV adds a query-side degree of freedom, enabling new attention directions. \\

$T_6$ & \textbf{Prefix query + key.} Both sides of the similarity come from the same learnable vector, but the output is still built from content values. & \textbf{Not representable in PT.} ATV can simultaneously steer queries and keys while still reading content values, providing a finer redistribution of attention mass that PT cannot mimic. \\

$T_7$ & \textbf{Prefix query + value.} Ordinary keys choose the weights; the returned information comes from a prefix-generated value. & \textbf{Not representable in PT.} PT can supply prefix values but cannot adapt the query; ATV adds this missing query modulation, enhancing expressivity. \\

$T_8$ & \textbf{Full prefix triad.} Query, key, and value are all produced by the same low-rank adapter, yielding a fully synthetic attention path. & \textbf{Not representable in PT.} PT has no mechanism for a fully synthetic attention channel without real tokens; ATV introduces an entirely new path, further enlarging the representational space. \\

\bottomrule
\end{tabular}
\end{table}

\paragraph{Key points of the theorem}

\begin{itemize}
    \item \textbf{Containment:}
    PT spans only the subspace generated by $T_1 + T_2$. ATV keeps those terms and introduces $T_3-T_8$, hence
    \[
    \mathcal{F}(\text{Attn}_{\text{prefix}}) \subseteq \mathcal{F}(\text{Attn}_{\text{ATV}})
    \]
    \item \textbf{Query-side freedom ($T_5,T_6,T_7,T_8$):}
    Because PT never changes $W_q$, any behavior that requires altering the query vector is strictly outside its representational span. ATV realizes this through the additive query $P^{\prime}_q$.
    \item \textbf{Mixed interactions ($T_3,T_4$):} 
    Unlike PT, ATV can blend a single soft prefix key or value with the untouched content tokens.
    To even approximate $T_3$, PT would have to add one custom prefix key for every content token, which is an impractical workaround, and $T_4$ cannot be reproduced by PT at all.
    \item \textbf{Full prefix channel ($T_8$):}
    A complete synthetic path lets ATV add task-specific information even when the original context is irrelevant, while still using no extra tokens at runtime.
\end{itemize}
Taken together, the additional six terms explain why ATV is more expressive: it augments the attention operator along every axis (query, key, and value) without introducing heavy retraining or large prefix matrices, yet it can still emulate PT as a special case.

\section{Detailed Experiments Setting}
\label{app:exp_setting}

Our experimental setup follows ELICIT~\cite{wang2025elicit}, using the same datasets and evaluation protocols. Below we provide detailed specifications.

\subsection{Dataset List}
\label{app:dataset-list}

All experiments are conducted on the same 20 in-domain tasks and 5 unseen tasks as used in ELICIT. Tasks are categorized as follows:

\begin{itemize}
    \item \textbf{Knowledge}: CommonsenseQA~\cite{talmor2018commonsenseqa}, OpenBookQA~\cite{mihaylov2018can}, HellaSwag~\cite{zellers2019hellaswag}, BoolQ~\cite{clark2019boolq}
    \item \textbf{Reasoning}: Four subsets from Big-Bench Hard (BBH)~\cite{suzgun2022challenging} (BBH Boolean Expressions, BBH Date Understanding, BBH Reasoning about Colored Objects, BBH Temporal Sequences), ARC-Challenge~\cite{clark2018think}
    \item \textbf{Mathematics}: MathQA~\cite{amini2019mathqa}, MMLU Pro-MATH~\cite{wang2024mmlu}
    \item \textbf{Safety}: Crows-Pairs~\cite{nangia2020crows}, BBQ-Age~\cite{parrish2021bbq}, Ethics-Commonsense, Ethics-Justice~\cite{merity2016pointer}
    \item \textbf{Natural Language Understanding (NLU)}: GLUE (SST-2, QNLI, MNLI)~\cite{wang2018glue}, SuperGLUE (WIC, RTE)~\cite{wang2019superglue}
    \item \textbf{Unseen}: GLUE COLA, BBQ-Religion, Deepmind~\cite{saxton2019analysing}, MMLU High School Psychology, BBH Logical Deduction Five objects
\end{itemize}

\subsection{Implementation Details and Baseline Configurations}

\paragraph{Common Setup.}
All experiments are conducted on NVIDIA A100 80GB GPUs. We use the same training data splits and evaluation protocols across all methods for fair comparison. Each experiment is repeated 3 times with different random seeds (42, 100, 10) to compute statistical significance. For training, we sample \textbf{90 examples per task} from the official training split of each in-domain task (excluding unseen tasks), and use the same sampled data across all baselines.

\paragraph{ATV.}
We train the small model \( \mathcal{M}_{\text{small}} \) (GPT-2, 137M parameters) and the expansion module \( f_\theta \) jointly with the following hyperparameters. A constant learning rate of \textbf{5e-4} is used without warmup or learning rate scheduling, along with \textbf{weight decay of 1e-5}. The model is optimized for \textbf{15 epochs} using the Adam optimizer. 

We inject a task vector \( v_{\text{ATV}} \in \mathbb{R}^{L \times d_l} \) into the last token’s hidden state at each layer as \( \tilde{h}^l = h^l + \lambda v_{\text{ATV}}^l \). 

We use a scaling factor of \( \boldsymbol{\lambda} = \mathbf{0.001} \) throughout all experiments. In our implementation, the hidden size of the small model is \( d_s = 768 \) (GPT-2), and the large models (LLaMA3-8B and Mistral-7B) use \( d_l = 4096 \) with \( L = 32 \) transformer layers.

\paragraph{ELICIT.}
We follow the official implementation and configuration of ELICIT. Task vectors are retrieved from a precomputed capability library, each paired with its optimal injection layer. At inference time, the selected vector is additively injected into the frozen LLM at the designated layer. All training and evaluation use the official codebase and default settings.

Each task vector is constructed from 10 exemplars per task, each with a 16-shot prompt. While the total number of unique samples may vary due to overlap, our analysis confirms a minimum of 91 unique samples per task. To ensure fair comparison, we use 90 training samples per task for all other baselines.

\paragraph{I2CL.}
We adopt the official I2CL implementation~\cite{li2025implicit}, modifying only the number of training epochs to 15 for consistency with other baselines. To ensure fair comparison, we deviate from the original setting, which calibrates context vectors and injection coefficients separately for each dataset using task identity. Instead, we train a shared set of coefficients across all datasets while keeping dataset-specific context vectors.

For evaluation on unseen tasks, we use a retrieval strategy that selects the most similar context vector among those obtained from in-domain datasets, based on cosine similarity between the input query and training prompts.

\paragraph{LoRA.}
We adopt the LoRA configuration described in the I2CL paper, which applies low-rank adaptation to the query and value projection matrices in all attention layers. The setup uses rank $r = 8$, scaling factor $\alpha = 32$, and a dropout rate of 0.05. All other settings, including the optimizer, follow the official implementation. 
However, as the original learning rate of $1\mathrm{e}{-3}$ resulted in poor performance in our setting, we adjust it to $4\mathrm{e}{-4}$.

\subsection{Task-Specific Prompt List}
\label{app:prompt_list}

We follow the prompt template settings from ELICIT, which adopts task-specific templates manually crafted based on guidelines from \texttt{lm-harness}~\cite{lintang_sutawika_2023_10256836} and the \texttt{chain-of-thought-hub}\footnote{\url{https://github.com/FranxYao/chain-of-thought-hub}}.
For each task, we use the same three distinct question templates as provided in ELICIT. The full set of question templates used for each task is listed in Table~\ref{tab:prompt_templates}.

The answer-side format is consistent across all tasks and composed of the following structure:
\begin{itemize}
    \item A line break (\texttt{\textbackslash n}) after the question template,
    \item A list of options in the form: \texttt{Options: (A) ..., (B) ..., (C) ..., ...},
    \item One of the following three answer prefixes:
    \begin{itemize}
        \item \texttt{A:}
        \item \texttt{Answer:}
        \item \texttt{The answer is}
    \end{itemize}
\end{itemize}

By combining the 3 question templates with the 3 answer prefixes, we construct 9 distinct prompt variants per task. Following the ELICIT setup, only the \texttt{A:} answer prefix is used during training, while all 3 answer formats are used during evaluation to assess generalization to unseen answer styles. This setting is consistently applied across all baseline methods.

\begin{longtable}{%
  >{\raggedright\arraybackslash}p{\dimexpr (4\textwidth - 24\tabcolsep)/17\relax}% Task (Dataset) column
  >{\raggedright\arraybackslash}p{\dimexpr (13\textwidth - 78\tabcolsep)/17\relax}% Template column (increased width)
}
\caption{\textbf{Question-side templates used for each task.} Each task uses three distinct prompt formats as provided in the original ELICIT setting.} \label{tab:prompt_templates} \\
\toprule
\textbf{Task (Dataset)} & \textbf{Template} \\
\midrule
\endfirsthead % This header appears on the first page

\multicolumn{2}{c}%
{{\tablename\ \thetable{} -- continued from previous page}} \\[0.5em]
\toprule
\textbf{Task (Dataset)} & \textbf{Template} \\
\midrule
\endhead % This header appears on subsequent pages

\midrule
\multicolumn{2}{r}{{Continued on next page}} \\
\endfoot % This footer appears on all pages except the last

\bottomrule
\endlastfoot % This footer appears on the last page

% Knowledge
\vspace{1.5ex}CommonsenseQA & \begin{itemize}[leftmargin=*,nosep]
    \item The following are multiple choice questions (with answers) about commonsense knowledge reasoning. Finish your answer with 'X' where X is the correct letter choice.\textbackslash n\textbackslash nQuestion: \{input\}
    \item Below are multiple-choice questions about commonsense reasoning. Answer with 'X', X being the correct option.\textbackslash n\textbackslash nQuestion: \{input\}
    \item Respond to these multiple-choice questions on commonsense knowledge. Conclude with 'X', where X is the right letter choice.\textbackslash n\textbackslash nQuestion: \{input\}
  \end{itemize} \\
\cmidrule{1-2}
\vspace{1.5ex}OpenBookQA & \begin{itemize}[leftmargin=*,nosep]
    \item The following are multiple choice questions (with answers) about multi-step reasoning. Finish your answer with 'X' where X is the correct letter choice.\textbackslash n\textbackslash nQuestion: \{input\}
    \item The following are multiple-choice questions testing multi-step reasoning. Answer with 'X', X being the correct option.\textbackslash n\textbackslash nQuestion: \{input\}
    \item Answer these multiple-choice questions involving multi-step logical thinking. Conclude with 'X', where X is the right letter choice.\textbackslash n\textbackslash nQuestion: \{input\}
  \end{itemize} \\
%\cmidrule{1-2}
\vspace{1.5ex}HellaSwag & \begin{itemize}[leftmargin=*,nosep]
    \item The following are multiple choice questions (with answers) about commonsense NLI. Finish your answer with 'X' where X is the correct letter choice.\textbackslash n\textbackslash nQuestion: \{input\}
    \item The following are multiple-choice questions about commonsense natural language inference. Answer with 'X', X being the correct option.\textbackslash n\textbackslash nQuestion: \{input\}
    \item Answer these multiple-choice questions on commonsense language understanding. Conclude with 'X', where X is the right letter choice.\textbackslash n\textbackslash nQuestion: \{input\}
  \end{itemize} \\
\cmidrule{1-2}
\vspace{1.5ex}BoolQ & \begin{itemize}[leftmargin=*,nosep]
    \item \{input\} \textbackslash nAnswer True or False.
    \item \{input\} \textbackslash nRespond with True or False.
    \item \{input\} \textbackslash nIs this statement correct? Answer True or False.
  \end{itemize} \\
\midrule

% Reasoning
\vspace{1.5ex}BBH Date Understanding & \begin{itemize}[leftmargin=*,nosep]
    \item Infer the date from context. Finish your answer with 'X' where X is the correct letter choice.\textbackslash n\textbackslash nQuestion: \{input\}
    \item Determine the date based on contextual clues. End your response with 'X', where X represents the correct option.\textbackslash n\textbackslash nQuestion: \{input\}
    \item Use the given context to deduce the date. Conclude your answer with 'X', X being the right letter choice.\textbackslash n\textbackslash nQuestion: \{input\}
  \end{itemize} \\
\cmidrule{1-2}
\vspace{1.5ex}BBH Boolean Expressions & \begin{itemize}[leftmargin=*,nosep]
    \item Evaluate the result of a random Boolean expression.\textbackslash n\textbackslash nQuestion: \{input\}
    \item Calculate the outcome of a given Boolean expression.\textbackslash n\textbackslash nQuestion: \{input\}
    \item Determine the result of the provided Boolean logic statement.\textbackslash n\textbackslash nQuestion: \{input\}
  \end{itemize} \\
\cmidrule{1-2}
\vspace{1.5ex}BBH Temporal Sequences & \begin{itemize}[leftmargin=*,nosep]
    \item Answer questions about which times certain events could have occurred. Finish your answer with 'X' where X is the correct letter choice.\textbackslash n\textbackslash nQ: \{input\}
    \item Determine possible occurrence times for specific events. Answer with 'X', X being the correct option.\textbackslash n\textbackslash nQ: \{input\}
    \item Identify when certain events could have happened. Conclude with 'X', where X is the right letter choice.\textbackslash n\textbackslash nQ: \{input\}
  \end{itemize} \\
\cmidrule{1-2}
\vspace{1.5ex}BBH Reasoning about Colored Objects & \begin{itemize}[leftmargin=*,nosep]
    \item Answer extremely simple questions about the colors of objects on a surface. Finish your answer with 'X' where X is the correct letter choice.\textbackslash n\textbackslash nQ: \{input\}
    \item Respond to basic questions about object colors on a surface. Answer with 'X', X being the correct option.\textbackslash n\textbackslash nQ: \{input\}
    \item Address simple queries regarding the colors of items on a surface. Conclude with 'X', where X is the right letter choice.\textbackslash n\textbackslash nQ: \{input\}
  \end{itemize} \\
%\cmidrule{1-2}
\vspace{1.5ex}ARC Challenge & \begin{itemize}[leftmargin=*,nosep]
    \item The following are multiple choice questions (with answers) about science. Finish your answer with 'X' where X is the correct letter choice.\textbackslash n\textbackslash nQuestion: \{input\}
    \item Below are multiple-choice science questions. Answer with 'X', X being the correct option.\textbackslash n\textbackslash nQuestion: \{input\}
    \item Respond to these multiple-choice questions on scientific topics. Conclude with 'X', where X is the right letter choice.\textbackslash n\textbackslash nQuestion: \{input\}
  \end{itemize} \\
\midrule

% Mathematics
\vspace{1.5ex}MathQA & \begin{itemize}[leftmargin=*,nosep]
    \item The following are multiple choice questions (with answers) about math word problem. Finish your answer with 'X' where X is the correct letter choice. \textbackslash n\textbackslash nQuestion: \{input\}
    \item Below are multiple-choice questions about math word problems. Answer with 'X', X being the correct option. \textbackslash n\textbackslash nQuestion: \{input\}
    \item Solve these multiple-choice math word problems. Conclude with 'X', where X is the right letter choice. \textbackslash n\textbackslash nQuestion: \{input\}
  \end{itemize} \\
\cmidrule{1-2}
\vspace{1.5ex}MMLU Pro-MATH & \begin{itemize}[leftmargin=*,nosep]
    \item The following are multiple choice questions (with answers) about math. Finish your answer with 'X' where X is the correct letter choice. \textbackslash n\textbackslash nQuestion: \{input\}
    \item Below are multiple-choice math questions. Conclude your response with 'X', X being the correct option. \textbackslash n\textbackslash nQuestion: \{input\}
    \item Answer these math multiple-choice questions. Answer with 'X', where X is the right letter choice. \textbackslash n\textbackslash nQuestion: \{input\}
  \end{itemize} \\
\midrule

% Safety
\vspace{1.5ex}Crows-Pairs & \begin{itemize}[leftmargin=*,nosep]
    \item \{input\} \textbackslash nQuestion: Which sentence is more stereotypical? Answer A or B.
    \item \{input\} \textbackslash nQuestion: Which sentence contains more stereotypical content? Answer A or B.
    \item \{input\} \textbackslash nQuestion: Identify the sentence with stronger stereotypical elements. Respond with A or B.
  \end{itemize} \\
\cmidrule{1-2}
\vspace{1.5ex}BBQ-Age & \begin{itemize}[leftmargin=*,nosep]
    \item The following are multiple choice questions (with answers) about social bias on age. Finish your answer with 'X' where X is the correct letter choice.\textbackslash n\textbackslash nQuestion: \{input\}
    \item The following are multiple-choice questions about age-related social bias. Answer with 'X', X being the correct option.\textbackslash n\textbackslash nQuestion: \{input\}
    \item Answer these multiple-choice questions on social bias regarding age. Conclude with 'X', where X is the right letter choice.\textbackslash n\textbackslash nQuestion: \{input\}
  \end{itemize} \\
\cmidrule{1-2}
\vspace{1.5ex}Ethics-Commonsense & \begin{itemize}[leftmargin=*,nosep]
    \item \{input\} \textbackslash nQuestion: Is this wrong? Answer Yes or No.
    \item \{input\} \textbackslash nQuestion: Is this ethically incorrect? Respond with Yes or No.
    \item \{input\} \textbackslash nQuestion: Does this violate ethical norms? Answer Yes or No.
  \end{itemize} \\
%\cmidrule{1-2}
\vspace{1.5ex}Ethics-Justice & \begin{itemize}[leftmargin=*,nosep]
    \item \{input\} \textbackslash nQuestion: Would most people believe this reasonable or unreasonable to say? Answer Yes or No.
    \item \{input\} \textbackslash nQuestion: Would the majority consider this statement reasonable or unreasonable? Answer Yes or No.
    \item \{input\} \textbackslash nQuestion: Is this statement generally perceived as reasonable or unreasonable? Respond with Yes or No.
  \end{itemize} \\
\midrule

% NLU
\vspace{1.5ex}GLUE (SST-2) & \begin{itemize}[leftmargin=*,nosep]
    \item \{input\} \textbackslash nQuestion: Is this sentence positive or negative? Answer positive or negative.
    \item \{input\} \textbackslash nQuestion: Does this sentence convey a positive or negative sentiment? Answer positive or negative.
    \item \{input\} \textbackslash nQuestion: Is the tone of this sentence positive or negative? Respond with positive or negative.
  \end{itemize} \\
\cmidrule{1-2}
\vspace{1.5ex}GLUE (QNLI) & \begin{itemize}[leftmargin=*,nosep]
    \item \{input\} \textbackslash nQuestion: Does this response answer the question? Answer Yes or No.
    \item \{input\} \textbackslash nQuestion: Is the response answer to the question? Answer Yes or No.
    \item \{input\} \textbackslash nQuestion: Does the given response address the question? Respond with Yes or No.
  \end{itemize} \\
\cmidrule{1-2}
\vspace{1.5ex}GLUE (MNLI) & \begin{itemize}[leftmargin=*,nosep]
    \item \{input\} True, False or Neither?
    \item \{input\} Answer selecting from: True, False, or Neither?
    \item \{input\} Choose one as answer: True, False, or Neither?
  \end{itemize} \\
\cmidrule{1-2}
\vspace{1.5ex}SuperGLUE (WIC) & \begin{itemize}[leftmargin=*,nosep]
    \item \{input\} \textbackslash nQuestion: Is the word used in the same way in the two sentences above? Answer Yes or No.
    \item \{input\} \textbackslash nQuestion: Is the word used similarly in both sentences above? Respond with Yes or No.
    \item \{input\} \textbackslash nQuestion: Does the word have the same meaning in the two given sentences? Answer Yes or No.
  \end{itemize} \\
\cmidrule{1-2}
\vspace{1.5ex}SuperGLUE (RTE) & \begin{itemize}[leftmargin=*,nosep]
    \item \{input\} \textbackslash nQuestion: Is the hypothesis correct based on the premise? Answer True or False.
    \item \{input\} \textbackslash nQuestion: Based on the premise, is the hypothesis accurate? Respond with True or False.
    \item \{input\} \textbackslash nQuestion: Does the premise support the hypothesis? Answer True or False.
  \end{itemize} \\
\midrule

% Unseen
\vspace{1.5ex}GLUE (COLA) & \begin{itemize}[leftmargin=*,nosep]
    \item \{input\} \textbackslash nQuestion: Does this sentence make sense? Answer Yes or No.
    \item \{input\} \textbackslash nQuestion: Is this sentence logically coherent? Respond with Yes or No.
    \item \{input\} \textbackslash nQuestion: Evaluate if this sentence is meaningful. Reply with Yes or No.
  \end{itemize} \\
%\cmidrule{1-2}
\vspace{1.5ex}BBQ-Religion & \begin{itemize}[leftmargin=*,nosep]
    \item The following are multiple choice questions (with answers) about social bias on religion. Finish your answer with 'X' where X is the correct letter choice.\textbackslash n\textbackslash nQuestion: \{input\}
    \item Here are multiple-choice questions addressing social biases related to religion. Conclude your answer with 'X', X being the correct letter option.\textbackslash n\textbackslash nQuestion: \{input\}
    \item These questions explore social biases in the context of religion. End your response with 'X', where X represents the right letter choice.\textbackslash n\textbackslash nQuestion: \{input\}
  \end{itemize} \\
\cmidrule{1-2}
\vspace{1.5ex}Deepmind & \begin{itemize}[leftmargin=*,nosep]
    \item The following are multiple choice questions (with answers) about algebraic word problems. Finish your answer with 'X' where X is the correct letter choice.\textbackslash n\textbackslash nQuestion: \{input\}
    \item Below are multiple-choice questions testing algebraic word problem solving skills. Conclude your answer with 'X', X being the correct option letter.\textbackslash n\textbackslash nQuestion: \{input\}
    \item These questions assess your ability to solve algebraic word problems. End your response with 'X', where X is the letter of the right choice.\textbackslash n\textbackslash nQuestion: \{input\}
  \end{itemize} \\
\cmidrule{1-2}
\vspace{1.5ex}MMLU High School Psychology & \begin{itemize}[leftmargin=*,nosep]
    \item The following are multiple choice questions (with answers) about high school psychology. Finish your answer with 'X' where X is the correct letter choice.\textbackslash n\textbackslash nQuestion: \{input\}
    \item Below are multiple-choice questions testing high school level psychology knowledge. Conclude your response with 'X', X representing the correct option.\textbackslash n\textbackslash nQuestion: \{input\}
    \item These questions assess understanding of high school psychology concepts. End your answer with 'X', where X is the letter of the correct choice.\textbackslash n\textbackslash nQuestion: \{input\}
  \end{itemize} \\
\cmidrule{1-2}
\vspace{1.5ex}BBH Logical Deduction Five Objects & \begin{itemize}[leftmargin=*,nosep]
    \item A logical deduction task which requires deducing the order of a sequence of objects. Finish your answer with 'X' where X is the correct letter choice.\textbackslash n\textbackslash nQuestion: \{input\}
    \item This challenge involves logically determining the sequence of a set of objects. Conclude your response with 'X', where X is the appropriate letter option.\textbackslash n\textbackslash nQuestion: \{input\}
    \item In this logical reasoning exercise, deduce the correct order of a series of objects. End your answer with 'X', X being the right letter choice.\textbackslash n\textbackslash nQuestion: \{input\}
  \end{itemize} \\
\bottomrule
\end{longtable}

\end{document}